\documentclass[lettersize,journal]{IEEEtran}
\usepackage{bm}
\usepackage{amsmath,amsfonts}
\usepackage{algorithmic}
\usepackage{algorithm}
\usepackage{array}

\usepackage{textcomp}
\usepackage{stfloats}
\usepackage{url}
\usepackage{verbatim}
\usepackage{graphicx}
\usepackage{cite}

\usepackage{threeparttable}

\usepackage{subfigure}

\usepackage{cite}
\usepackage{amsmath,amssymb,amsfonts}
\usepackage{algorithmic}
\usepackage{graphicx}
\usepackage{textcomp}
\usepackage{caption}
\usepackage{subcaption}
\usepackage{algorithm}
\usepackage{booktabs}
\usepackage{tabularx}
\newcolumntype{Y}{>{\centering\arraybackslash}X}
\usepackage{hyperref}

\usepackage{rotating}
\usepackage{ragged2e} 
\usepackage{color} 
\usepackage{adjustbox}
\usepackage{makecell}
\usepackage{threeparttable}
\usepackage{multirow}
\usepackage[super]{nth}

\usepackage[table]{xcolor}
\usepackage{colortbl}       

\makeatother
\begin{document}

\title{A Survey on Deep Multi-Task Learning in Connected Autonomous Vehicles}

\author{Jiayuan Wang, \IEEEmembership{Graduate Student Member,~IEEE}, Farhad Pourpanah, \IEEEmembership{Senior Member,~IEEE}, \\ Q. M. Jonathan Wu, \IEEEmembership{Senior Member,~IEEE}, and Ning Zhang, \IEEEmembership{Senior Member,~IEEE}
\thanks{This research was undertaken, in part, thanks to funding from the Canada Research Chairs Program, and in part by the NSERC’s CREATE program on TrustCAV. (Corresponding author: Q. M. Jonathan Wu; Ning Zhang.)}
\thanks{Jiayuan Wang, Q. M. Jonathan Wu and Ning Zhang are with the Department of Electrical and Computer Engineering, University of Windsor, Windsor, ON N9B 3P4, Canada (e-mails: wang621@uwindsor.ca, jwu@uwindsor.ca and ning.zhang@uwindsor.ca)}

\thanks{Farhad Pourpanah is with the Department of Electrical and Computer Engineering, Queen’s University, Kingston, ON, Canada (e-mail: farhad.086@gmail.com)}

}

\maketitle

\begin{abstract}
Connected autonomous vehicles (CAVs) must simultaneously perform multiple tasks, such as perception, prediction, planning, and control, to ensure safe and reliable navigation in complex environments. Moreover, through vehicle-to-everything (V2X) communication, cooperative perception and driving among CAVs can be enabled, thereby mitigating the limitations of individual vehicles, while it also introduces stringent latency, reliability, and bandwidth constraints. Traditionally, tasks are addressed using separate models, which leads to high deployment costs, increased computational overhead, and challenges in achieving real-time performance. Multi-task learning (MTL) has recently emerged as a promising solution that enables the joint learning of multiple tasks within a unified model. This offers improved efficiency and resource utilization. To the best of our knowledge, this survey is the first comprehensive review focusing on deep MTL in CAVs. We begin with an overview of CAVs and MTL to provide foundational background. Then, we review MTL approaches across key functional domains in CAVs, including perception, prediction, planning, control, as well as V2X communications and radio resource management (RRM). For the first four domains, we categorize existing works under ego vehicle-only (onboard-only) and V2X-enhanced cooperative (multi-agent) paradigms. We further discuss V2X communications and RRM as communication-centric MTL problems. Finally, we discuss the strengths and limitations of existing methods, identify key research gaps, and provide future research directions aimed at advancing MTL methodologies for CAV systems.
\end{abstract}

\begin{IEEEkeywords}
Multi-task learning, connected autonomous vehicles, deep learning, V2X communications
\end{IEEEkeywords}

\section{Introduction}
\label{sec:introduction}

\IEEEPARstart{C}{onnected} autonomous vehicles (CAVs) combine autonomous driving capabilities with vehicle-to-everything (V2X) communication \cite{hafner2022survey} and enable information exchange with other vehicles, roadside infrastructure, and cloud servers \cite{wang2020architectural}. This connectivity supports cooperative decision-making beyond the line of sight of onboard sensors, thereby enhancing environmental awareness, safety, and traffic management \cite{gu2025digital}. Specifically, through cooperative perception, V2X communication complements onboard sensing by providing a more comprehensive understanding of the driving environment \cite{garcia2021tutorial, balkus2022survey, katare2023survey}. This extended situational awareness is effective in mitigating the limitations of onboard sensors, especially under occlusions or at long distances. Such benefits are especially pronounced in densely populated or complex urban scenarios, where visibility and sensing coverage are often constrained \cite{chu2025occlusion, bae2025rethinking, wang2019networking}.

As a foundation of CAVs, autonomous driving capabilities are supported by autonomous driving systems (ADS), which must simultaneously execute multiple tasks, e.g., lane segmentation, object detection, distances and trajectory estimation, and real-time longitudinal and lateral control (i.e., throttle/braking and steering), to enable safe navigation in dynamic and complex environments \cite{hussain2019autonomous,behere2016functional}. 
Current ADS operate either through structured pipelines composed of perception, prediction, planning, and control modules \cite{yurtsever2020survey} or through end-to-end approaches that directly map sensor inputs to control commands. Modular pipelines remain the dominant paradigm due to their interpretability and compatibility with safety constraints. These modules integrate diverse sensor information to enable real-time decisions in response to the surrounding environment. 
The core tasks in these modules, e.g., object detection \cite{mahaur2023small}, semantic segmentation \cite{rossolini2024real}, trajectory prediction \cite{yang2024multi, wu2023language}, depth estimation \cite{ishihara2021multi}, motion prediction \cite{rao2024enhancing, yang2018end}, and behavior prediction \cite{baicang2025multi}, have been addressed independently, i.e., each task requires its distinct model and architecture. Although this strategy has achieved considerable success, it suffers from high model development costs, substantial computational demands, and difficulties in meeting real-time performance requirements \cite{wang2024you, teichmann2018multinet}.

Environmental perception in ADS can be categorized into multi-sensor fusion \cite{liu2022robust, xiao2022multimodal} and camera-only methods \cite{wang2024you, wu2022yolop, wang2023sparse}. Multi-sensor fusion techniques utilize data from sensors such as LiDAR, radar, and cameras to construct more comprehensive scene representations. For example, Waymo driver perception system \cite{waymo_driver} integrates LiDAR, cameras, and radar data to obtain environmental information. However, these methods involve high costs and increased system complexity \cite{de2021evaluating, li2023emergent}. In contrast, camera-only methods offer a cost-effective alternative with simpler integration requirements \cite{broggi2013extensive}. By combining deep learning (DL) models with camera-based methods, rich visual information can be captured to perform perception tasks. As a result, camera-based ADS have become attractive due to their scalability, affordability, and potential for widespread adoption \cite{carranza2020performance}. Tesla Vision is one such system, i.e., camera-only autopilot, reported to achieve comparable or improved active safety ratings and superior pedestrian automatic emergency braking performance relative to traditional radar-based systems \cite{TeslaVision2024}. 

Recent advances in computer vision have accelerated the development of ADS. These systems now demonstrate improved capabilities in understanding dynamic road environments and reliably interpreting changing traffic scenes and road conditions. Such progress holds the potential to transform transportation by enhancing urban mobility, optimizing logistics efficiency, and enabling smarter traffic management.  However, realizing these benefits in practice requires an ADS to run an increasing set of tasks simultaneously under strict real-time and onboard compute constraints. Deploying separate task-specific models leads to high computational demands and model development costs \cite{wang2024you, teichmann2018multinet}. To address these challenges, multi-task learning (MTL) offers a promising solution by integrating multiple tasks within a single unified model. MTL leverages shared computational components to improve efficiency and facilitate real-time performance, which is critical for ADS. Furthermore, the integration of multiple tasks within one model contributes to more robust and reliable predictions \cite{liang2022effective}. 
MTL \cite{caruana1997multitask} enhances generalization through inductive transfer by leveraging domain information from task-specific data as an inductive bias. This is achieved by parallel learning of multiple tasks, where a shared representation enables knowledge from one task to support others. 

\newcolumntype{P}[1]{>{\raggedright\arraybackslash}p{#1}}

\begin{table*}[t]

\centering
\caption{Summary of related surveys.}
\label{tab:related_survey_summary_ads_v2x_comm_mtl}

\footnotesize
\setlength{\tabcolsep}{2.0pt}
\renewcommand{\arraystretch}{1.12}

\resizebox{\textwidth}{!}{
\begin{tabular}{P{0.06\textwidth} c p{0.56\textwidth} c c c c c c c}
\toprule
Ref. & Year & Description & Perception & Prediction & Planning & Control & V2X-C & Com & MTL \\
\midrule

\cite{vandenhende2022multi} & 2022 &
Surveys deep MTL for dense prediction by reviewing architectures and optimization methods, and presenting extensive evaluations across dense prediction benchmarks. &
$\times$ & $\times$ & $\times$ & $\times$ & $\times$ & $\times$ & \checkmark \\
\addlinespace

\cite{zhang2022surveyzhang} & 2022 &
Surveys MTL from algorithmic modeling, applications, and theoretical analyses, and categorizes supervised MTL into five approaches. &
$\times$ & $\times$ & $\times$ & $\times$ & $\times$ & $\times$ & \checkmark \\
\addlinespace

\cite{hafner2022survey} & 2022 &
Surveys V2X-enabled cooperative maneuvers by classifying cooperation architectures, comparing coordination protocols, and summarizing relevant standards and projects. &
$\times$ & $\times$ & \checkmark & \checkmark & \checkmark & \checkmark & $\times$ \\
\addlinespace

\cite{islam2023connected} & 2023 &
Reviews CAV practice by surveying autonomy technologies, V2X communications, computation needs, infrastructure readiness, testbeds, and deployment challenges. &
\checkmark & $\times$ & \checkmark & \checkmark & \checkmark & \checkmark & $\times$ \\
\addlinespace

\cite{zhao2023multi} & 2023 &
Reviews deep MTL for medical imaging by summarizing cascaded, parallel, interacted, and hybrid architectures, surveying applications across anatomical regions. &
$\times$ & $\times$ & $\times$ & $\times$ & $\times$ & $\times$ & \checkmark \\
\addlinespace

\cite{fang2024anomaly} & 2024 &
Surveys anomaly diagnosis in CAVs by covering detection and interpretation across sensing, communications, planning, and control from a safety and security perspective. &
\checkmark & $\times$ & \checkmark & \checkmark & \checkmark & \checkmark & $\times$ \\
\addlinespace

\cite{alzahrani2024survey} & 2024 &
Surveys MTL in smart transportation and categorizes applications such as traffic forecasting, traffic sign recognition, taxi demand prediction, and autonomous driving. &
\checkmark & \checkmark & $\times$ & \checkmark & $\times$ & $\times$ & \checkmark \\
\addlinespace

\cite{zhang2025vehicle} & 2025 &
Surveys V2X communication for intelligent connected vehicles, analyzing cooperative communication phases, scenarios, and datasets with evaluation practices. &
\checkmark & \checkmark & \checkmark & \checkmark & \checkmark & \checkmark & $\times$ \\
\addlinespace

\cite{liu2025collaborative} & 2025 &
Reviews collaborative perception in intelligent connected vehicles, covering mechanisms, collaboration strategies, evaluation, security, and applications. &
\checkmark & $\times$ & $\times$ & $\times$ & \checkmark & \checkmark & $\times$ \\
\addlinespace

\cite{huang2025vehicle} & 2025 &
Surveys V2X cooperative perception for autonomous driving using a pipeline taxonomy, and summarizes constraints, datasets, evaluation, and real-world implementations. &
$\checkmark$ & $\times$ & $\times$ & $\times$ & $\checkmark$ & $\checkmark$ & $\times$ \\
\addlinespace

\cite{wang2025review} & 2025 &
Reviews vision-based multi-task perception for autonomous driving, covering key tasks and summarizing MTL definitions, architectures, loss design, datasets, and metrics. &
\checkmark & $\times$ & $\times$ & $\times$ & $\times$ & $\times$ & \checkmark \\
\addlinespace

\textbf{Our survey} &  &
Review of deep MTL for CAVs across perception, prediction, planning, and control under ego vehicle-only and V2X-enhanced cooperative paradigms, as well as communication-centric MTL studies for V2X communications and RRM. &
\checkmark & \checkmark & \checkmark & \checkmark & \checkmark & \checkmark & \checkmark \\
\bottomrule
\end{tabular}}

\vspace{2pt}
\footnotesize
\raggedright
\justifying
\textit{Note:}
\textbf{V2X-C}: V2X-enhanced cooperation.
\textbf{Com}: Communication or radio resource management objectives. \checkmark indicates dedicated discussions or taxonomy of the module beyond high-level mentions or block-diagram illustrations; otherwise it is marked as $\times$.
\end{table*}

In CAVs, MTL enables the integration of tasks such as object detection, semantic segmentation, lane detection, and drivable area segmentation within a single model \cite{wang2024rtmdet, wang2024you, zhan2024yolopx}. MTL models can jointly analyze multiple tasks within the driving environment, resulting in more comprehensive predictions \cite{ishihara2021multi, wu2022yolop}. For example, the segmentation mask can provide spatial priors for object detection, while bounding boxes and category information from detection tasks can inform semantic segmentation \cite{zhang2021loss}. Training across multiple tasks also enables the use of diverse data sources, thereby increasing model robustness, reducing overfitting, and improving overall performance \cite{zhang2022surveyzhang}. It can also reduce the number of required training samples, which is particularly valuable in CAV applications since annotating data such as bounding boxes and segmentation masks is labor-intensive and time-consuming. While semi-supervised \cite{li2025semi, yang2025unimatch}, self-supervised \cite{liu2025comprehensive, radwan2025tutorial}, and unsupervised \cite{nisioti2018intrusion} methods have been explored to alleviate annotation cost, MTL improves data efficiency from a different perspective by enabling related tasks to share representations. Specifically, the inductive bias from other tasks helps each task generalize better and allows each task to avoid overfitting with fewer labeled samples. Additionally, MTL improves computational efficiency by integrating multiple tasks into a single model. This reduces system complexity and lowers the need for computing resources, compared to training and maintaining separate models for each task. For example, the study in \cite{wang2024you} shows that handling the three tasks individually requires two YOLOv8n(seg) models for drivable area and lane line segmentation (each with 3.26M parameters) and one YOLOv8n(det) model for object detection (3.16M parameters), resulting in a total of 9.68M parameters. In contrast, the corresponding MTL model, A-YOLOM(n), requires only 4.43M parameters while achieving better overall performance across all tasks. This efficiency is especially critical for real-time CAV applications, where timely and accurate decision-making within the limited computing resources is essential for safety and operational performance \cite{wu2017squeezedet, weng2018pole, lin2018architectural}. However, not all tasks work well together. Sharing features between unrelated tasks can degrade the performance of all tasks. This issue is known as the negative transfer phenomenon \cite{ruder2017overview}. Therefore, it is essential to carefully assess task relationships and consider the risk of negative transfer when designing MTL architectures. Moreover, MTL can exhibit the seesaw phenomenon, where improving the performance of one task may lead to degradation in others \cite{tang2020progressive}.

\begin{figure}[t]
    \centering
    \includegraphics[width=0.98\linewidth]{image/structure_of_org.pdf}
    \caption{Organization of this paper.}
    \label{fig:organization_of_this_paper}
\end{figure}

\subsection{Contributions}
Over the years, many surveys have been conducted on MTL and CAVs from both methodological and system-level perspectives. On the MTL side, existing surveys have reviewed MTL methods across a wide range of domains and learning frameworks, including dense prediction in computer vision \cite{vandenhende2022multi}, medical image analysis \cite{zhao2023multi}, as well as general DL-oriented overviews \cite{zhang2022surveyzhang}.
However, these surveys have not examined how deep MTL is integrated within complex autonomous systems such as CAVs.

In parallel, several surveys have focused on CAV-related research from system-level and deployment perspectives, including collaborative sensing and communication pipelines \cite{liu2025collaborative}, V2X cooperative perception with a pipeline taxonomy \cite{huang2025vehicle}, cooperative architectures and maneuver coordination protocols \cite{hafner2022survey}, anomaly diagnosis and safety-oriented analysis \cite{fang2024anomaly}, and broader technology stacks and impacts of CAVs \cite{islam2023connected, zhang2025vehicle}.
These surveys primarily emphasize system architectures, communication technologies, safety considerations, and deployment considerations, while providing limited insight into learning mechanisms, particularly the role of deep MTL across CAV functions.

To date, only a very limited number of surveys have reviewed the intersection of deep MTL and autonomous driving. Existing works either treat autonomous driving as one application scenario within broader smart transportation MTL surveys or focus on specific ego vehicle perception tasks, and they do not systematically address V2X-enhanced cooperation \cite{alzahrani2024survey, wang2025review}.
For example, Alzahrani et al. \cite{alzahrani2024survey} surveyed MTL for smart transportation. However, its autonomous-driving discussion is mainly example-driven rather than a system-level review organized along the ADS software stack, and it does not analyze V2X-enhanced cooperative driving or communication and radio resource management (RRM) objectives.
By contrast, Wang et al. \cite{wang2025review} reviewed vision-based multi-task perception for autonomous vehicles, focusing on perception. However, it does not provide coverage across prediction, planning, and control, nor V2X-integrated scenarios.

To address this gap, this survey provides a comprehensive review of deep MTL methods in CAVs published between 2018 and 2025 across perception, prediction, planning, and control under both ego vehicle-only and V2X-enhanced cooperative paradigms. In addition, it summarizes communication-centric MTL designs for V2X communications and RRM, an aspect largely overlooked in prior work. 
Table \ref{tab:related_survey_summary_ads_v2x_comm_mtl} presents a comparison of our survey with related surveys, indicating their coverage of ADS functions, V2X-enhanced cooperation, and communication/RRM objectives, as well as whether MTL is considered. This comparison helps delineate the scope and contributions of this survey and highlights open challenges and future research directions in MTL-enabled CAV systems. 

The main contributions of this review include:
\begin{itemize}
    \item To the best of our knowledge, this survey is the first comprehensive review focused on MTL in CAVs, including an overview of CAVs and MTL as foundational background. Then, we bridge the MTL and CAV domains by discussing the motivations for applying MTL in CAVs.

    \item We propose a taxonomy that organizes deep MTL in CAVs by functional domain and operational context. The taxonomy distinguishes ego vehicle-only, V2X-enhanced cooperative, and communication-centric MTL problems, reflecting different system constraints, information availability, and architectural design choices.
    
    \item Based on the proposed taxonomy, we provide a systematic review of deep MTL methods in CAVs, covering both ego vehicle-only and V2X-enhanced cooperative paradigms across perception, prediction, planning, and control. We further summarize communication-centric, model-level MTL designs for V2X communications and RRM.

    \item We distill cross-domain insights and deployment-oriented considerations, and identify key research gaps and future directions toward deployable, real-time, and safety-critical MTL in CAVs.
\end{itemize}

{\color{black}
\subsection{Organization}
The rest of this paper is organized as follows. Section \ref{section2} provides a comprehensive review of CAV systems by categorizing them into three components: the hardware layer, the software layer, and V2X communication. Section \ref{section3} provides an overview of MTL, encompassing problem formulation, architectural paradigms, optimization strategies, and motivations for its application in CAVs with our proposed taxonomy of deep MTL in CAVs. Sections \ref{sec:mtl_perception} --\ref{sectionivd} are organized by functional domains (perception, prediction, planning and control, as well as V2X communications and RRM). Within the perception, prediction, and planning and control domains, both ego vehicle-only and V2X-enhanced cooperative MTL settings are discussed, while the V2X communications and RRM section focuses on communication-centric MTL problems. Section \ref{section5} provides a cross-domain discussion and distills the key research gaps and future directions. 
Section \ref{sec:conclusion} concludes the survey. 
Fig. \ref{fig:organization_of_this_paper} indicates the organization of this survey. 
Table \ref{table:abbreviations} summarizes abbreviations used in this paper.
}

\begin{table*}[t]
  \centering
  
  \renewcommand{\arraystretch}{1.2}
  \setlength{\tabcolsep}{12pt} 
  \caption{Summary of Important Abbreviations in Alphabetical Order.}
  \label{table:abbreviations}
  \begin{tabularx}{\textwidth}{l X l X}
    \hline
    \textbf{Abbr.} & \textbf{Definition} & \textbf{Abbr.} & \textbf{Definition} \\
    \hline
    ADS   & Autonomous driving systems & AoI   & Age of Information \\
    BEV   & Bird's-eye view & C-CNN & One-stage conventional CNN \\
    C-V2X & Cellular vehicle-to-everything & CAM   & Cooperative awareness message \\
    CAVs  & Connected autonomous vehicles & CLIP  & Contrastive language--image pre-training \\
    CNN   & Convolutional neural network & Depth & Depth estimation \\
    Det   & Object detection & DL    & Deep learning \\
    FOcc  & Future occupancy forecasting & FPS   & Frames per second \\
    GNN   & Graph neural network & GRU   & Gated recurrent unit \\
    H     & Hard-parameter sharing & HC-Trans & Hybrid CNN-Transformer architectures \\
    Hy    & Hybrid-parameter sharing & IoU   & Intersection over union \\
    KPI   & Key performance indicator & LSTM  & Long short-term memory \\
    Map   & HD/BEV map & Mot   & Motion forecasting \\
    MPC   & Model predictive controller & MTL   & Multi-task learning \\
    PDR   & Packet delivery rate & PID   & Proportional--integral--derivative \\
    R-CNN & Region-based convolutional neural network & RL    & Reinforcement learning \\
    RRM   & Radio resource management & RSU   & Roadside unit \\
    S     & Soft-parameter sharing & SemSeg & Semantic segmentation \\
    SNR   & Signal-to-noise ratio & SSD   & Single-shot multi-box detector \\
    Track & Tracking & TrajPred & Trajectory forecasting \\
    VLM   & Vision-language model & - & Not applicable.\\
    \hline
  \end{tabularx}

\end{table*}

\newlength{\subfigimgheight}
\setlength{\subfigimgheight}{3.5cm} 

\begin{figure*}[t]
  \centering

  \subfigure[Ego vehicle modular architecture for CAV, where sensor data are processed through perception, prediction, planning, and control modules, while V2X enables information exchange with nearby agents.]{
    \begin{minipage}[c]{0.28\textwidth}
      \centering
      \begin{minipage}[c][\subfigimgheight][c]{\linewidth}
        \centering
        \includegraphics[width=1\linewidth]{image/survey_2a.pdf}
      \end{minipage}
      \label{fig:subfig2a}
    \end{minipage}
  }
  \hfill
  \subfigure[Ego vehicle end-to-end architecture for CAV, where sensor data and V2X information are directly mapped to control commands through a unified framework.]{
    \begin{minipage}[c]{0.30\textwidth}
      \centering
      \begin{minipage}[c][\subfigimgheight][c]{\linewidth}
        \centering
        \includegraphics[width=1\linewidth]{image/survey_2b.pdf}
      \end{minipage}
      \label{fig:subfig2b}
    \end{minipage}
  }
  \hfill
  \subfigure[V2X-enabled cooperative CAV scenario, where information from nearby vehicles, pedestrians, and infrastructure augments onboard sensing to support decision-making.]{
    \begin{minipage}[c]{0.32\textwidth}
      \centering
      \begin{minipage}[c][\subfigimgheight][c]{\linewidth}
        \centering
        \includegraphics[width=1\linewidth]{image/CAVs.pdf}
      \end{minipage}
      \label{fig:subfigCAV}
    \end{minipage}
  }
  \vspace{-0.1 cm}
  \caption{Overview of CAVs.}
  \label{fig:overview_ADS}
\end{figure*}

\section{Overview of CAVs}
\label{section2}
CAVs build upon traditional ADS by integrating V2X communication capabilities (see Fig. \ref{fig:overview_ADS}). This connectivity enables cooperative perception and decision-making beyond the line of sight of onboard sensors. Figs. \ref{fig:subfig2a} and \ref{fig:subfig2b} illustrate two representative ego vehicle computing strategies for CAVs. Specifically, Fig. \ref{fig:subfig2a} shows a modular architecture, where the ego vehicle perceives the surrounding environment via onboard sensors, receives cooperative information from other agents through the V2X interface, and executes control commands through actuators within a structured pipeline consisting of perception, prediction, planning, and control. Fig. \ref{fig:subfig2b} shows an end-to-end architecture, where heterogeneous inputs from onboard sensing and V2X communication are mapped directly to control commands through an end-to-end framework. Fig. \ref{fig:subfigCAV} presents a detailed view of a V2X-enhanced cooperative scenario involving multiple vehicles, pedestrians, and infrastructure. In this setting, each CAV consists of a hardware layer, a software layer, and a V2X communication module. We discuss each component in the following subsections. 

\subsection{Hardware Layer}
This layer is responsible for perceiving the environment, executing control commands, and processing data in real-time. It includes all physical components, including sensors, computing platforms, and actuators.  

\subsubsection{Sensors}

Sensors in ADS can be broadly categorized into exteroceptive and proprioceptive, which play distinct yet complementary roles \cite{campbell2018sensor,ignatious2022overview}. Exteroceptive sensors (e.g., cameras, LiDAR, radar, and ultrasonic sensors) characterize the external environment by capturing appearance, geometry, and motion cues, whereas proprioceptive sensors (e.g., IMU and GNSS) measure vehicle ego states to support motion estimation and localization \cite{campbell2018sensor}. Exteroceptive sensing typically provides rich spatial context but is more sensitive to environmental conditions, while proprioceptive sensing offers high-rate measurements yet may suffer from drift or degraded global accuracy in challenging environments \cite{chen2024deep,liu2021computing}. In practice, ADS commonly fuse both sensor types to improve robustness and accuracy for perception and localization \cite{ignatious2022overview}.

\subsubsection{Computing platforms}

Modern vehicles typically incorporate multiple highly powerful central computers. Each computer is responsible for different computing domains. For example, Bosch \cite{BoschVehicleComputer} categorizes these domains into powertrain, chassis, driver assistance, and infotainment. ADS belong to the driver assistance domain, which requires real-time processing of extensive sensor data for tasks such as perception, sensor fusion, and decision-making. To meet these requirements, specialized computing platforms, such as NVIDIA Drive Orin \cite{nvidia_drive_agx}, Horizon Journey 6 \cite{horizon_journey_series}, Mobileye EyeQ6 \cite{mobileye_eyeq}, and Qualcomm Snapdragon Ride \cite{qualcomm_snapdragon_ride}, have been developed for driver assistance.

\subsubsection{Actuators} 
Actuators convert control module outputs into physical vehicle movements. Modern actuators replace traditional mechanical linkages with electronically controlled systems \cite{obertino2023upgrade, sahboun2022controller}, enabling faster, more precise, and stable responses for automated steering, acceleration, and braking. Main actuators include steering actuators, throttle actuators, brake actuators, and other actuators.

\subsection{Software Layer}
This layer is responsible for converting raw sensor inputs into actionable control commands, either through modular processing or an end-to-end framework. The modular architecture decomposes the pipeline into perception, prediction, planning, and control modules, which jointly interpret sensor data, forecast surrounding agents, generate safe trajectories, and issue actuation commands. In contrast, the end-to-end architecture integrates these functions within a unified framework rather than separating them into modules. The computing platform serves as the backbone for running these software functions with low latency and high reliability.

\subsubsection{Perception}
Perception serves as the ``eyes" of ADS. It is responsible for processing the sensor data to interpret the vehicle's surroundings. Core tasks include object detection (identifying and localizing vehicles, pedestrians, cyclists, etc.), semantic segmentation (identifying drivable areas, lane markings, sidewalks, etc.), instance segmentation (categorizing objects and distinguishing individual instances at the pixel level), traffic light and sign recognition, and localization (ego-vehicle position and orientation estimation). Some systems further employ multi-sensor fusion \cite{liu2022robust, xiao2022multimodal} to improve robustness under occlusion and adverse conditions.

\begin{figure}[t]
    \centering
    \includegraphics[width=\linewidth]{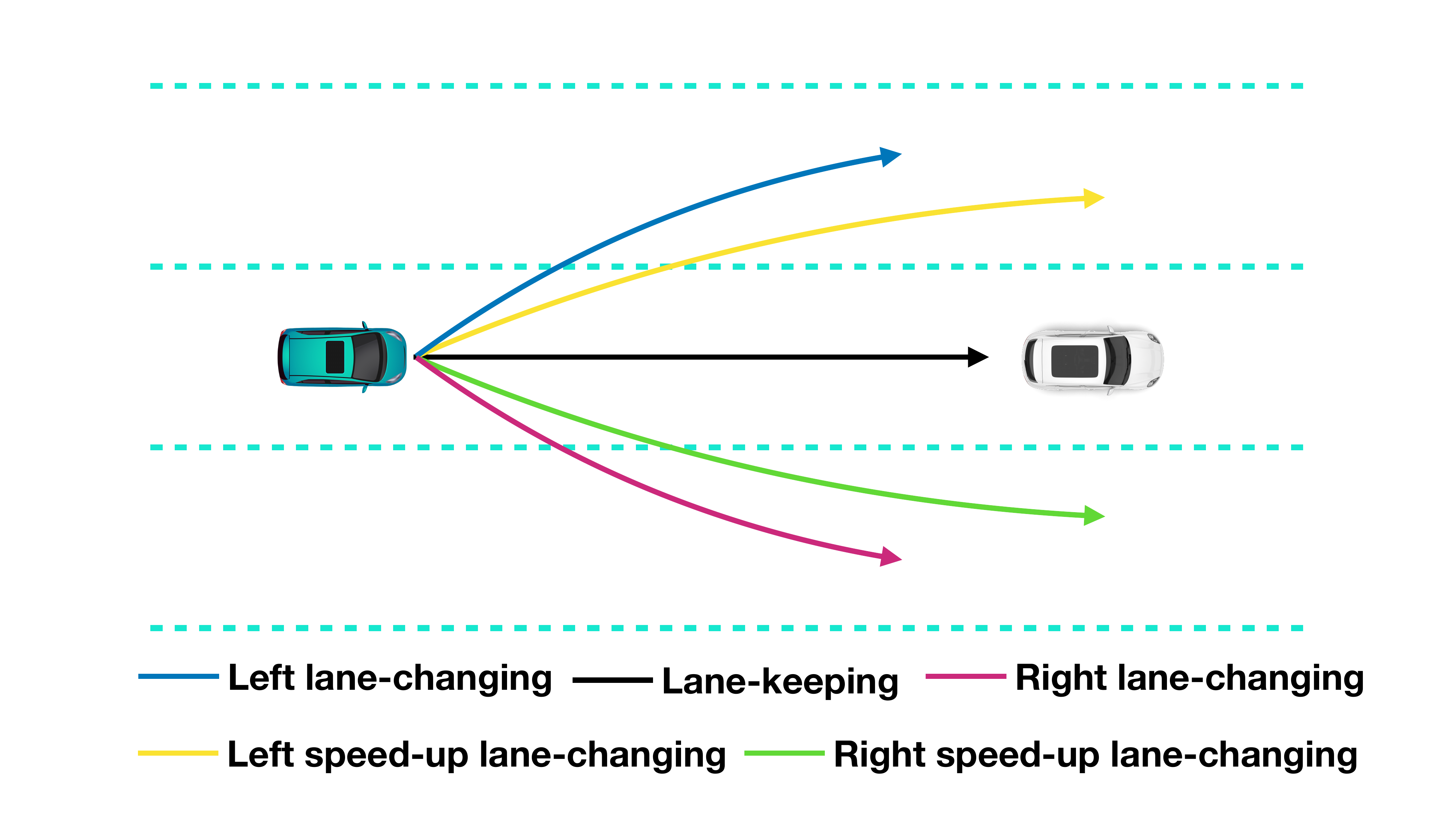}
    \caption{Driving behavior intention diagram (adapted from \cite{gao2024probabilistic}.)}
    \label{fig:intention}
\end{figure}

\subsubsection{Prediction} 
Prediction operates as a bridge between the perception and planning modules by predicting future states of surrounding agents. It predicts the future trajectories or behaviors of the other agents, such as vehicles and pedestrians. It involves time-series modeling and understanding of agent intent \cite{yang2024multi, yuan2024temporal}, as illustrated in Fig. \ref{fig:intention}. Key tasks include trajectory forecast (estimating the sequential future position of each agent) and classifying driving behavior intentions (turning, speed-up turning, or lane-keeping). In CAV scenarios, the ego vehicle can also obtain planning results shared by other agents through V2X communication. From the ego vehicle’s perspective, these shared planning results serve as high-confidence predictions. It further reduces uncertainty and enhances decision accuracy.

\subsubsection{Planning} 
Planning determines the ego vehicle's future path based on the output from the perception and prediction modules. It includes decision-making and motion planning. Decision-making comprises the scenario manager (identifying driving scenarios like highway cruising, intersection crossing, or lane changing), stage processing (breaking scenarios into discrete steps), and behavior decision (choosing actions like lane changes or stops). This decision-making supports the motion planning component in generating an executable trajectory according to vehicle dynamics, safety constraints, and comfort requirements. Traditionally, motion planning adopts sampling-based \cite{chen2025efficient, ogretmen2024sampling}, optimization-based \cite{jekl2025scenario, guo2024spatio}, or rule-based heuristics \cite{bouchard2022rule, ding2019rule} methods, all aiming to generate safe and efficient paths for the vehicle to follow.

\begin{table*}[t]
\centering
\caption{Comparison between ego vehicle-only and V2X-enhanced cooperative paradigms.}
\label{table:single-multi-agent}
\begin{tabular}{p{3.5cm} p{6cm} p{7cm}}
\toprule
\textbf{Feature} & \textbf{Ego vehicle-only paradigm} & \textbf{V2X-enhanced cooperative paradigm} \\
\midrule
Sensor Coverage & Limited to ego vehicle's field of view; suffers from occlusion & Expanded through shared sensing; mitigates blind spots \\
Information Completeness & Incomplete in complex or distant scenes & Access to extended scene context via V2X \\
Interaction Modeling & Relies on local inference; limited accuracy & Sharing motion state improves prediction \\
Robustness & Vulnerable to sensor failures and limited visibility & More robust via shared multi-view sensing \\
Communication Requirement & Fully self-contained; no networking needed & Requires reliable, low-latency communication infrastructure \\
Deployment Complexity & Low (easier to deploy and validate) & High (requires synchronization and supporting infrastructure) \\
Typical Scenarios & Highways, sparse traffic, structured roads & Urban intersections, merging lanes, occluded environments \\
\bottomrule
\end{tabular}
\end{table*}

\subsubsection{Control} \label{subsection:sc} 
Control converts the planned trajectory into low-level actuator commands, such as steering, throttle, and braking. These commands are executed by feedback controllers, where proportional-integral-derivative (PID) control \cite{zhai2025design, dong2021autonomous} and model predictive control (MPC) \cite{hu2025novel, xu2024parallel} are commonly used for trajectory tracking. The strategy that combines MPC and PID has also been explored to reduce steady-state tracking errors while enhancing steering smoothness and robustness to model simplifications \cite{chu2023trajectory}. Alternatively, end-to-end systems predict commands from raw sensor inputs using a single model \cite{kim2022end, yang2018end}. Compared to the modular pipelines, end-to-end approaches provide a unified framework that jointly optimizes perception, prediction, planning, and control \cite{yurtsever2020survey}, which simplifies architectures and improves computational efficiency. However, they also introduce some challenges related to interpretability, safety guarantees, and causal ambiguity \cite{chen2024end}. 

\subsection{V2X communication} 
V2X communication enables data exchange between vehicles and external entities, including vehicle-to-vehicle (V2V), vehicle-to-infrastructure (V2I), vehicle-to-pedestrian (V2P), and vehicle-to-network (V2N) interactions. By complementing onboard sensing with remote information, V2X extends perception beyond line-of-sight and sensor range limitations and supports cooperative driving in intelligent transportation systems \cite{ren2024interruption}. Through such information sharing, vehicles can obtain more comprehensive knowledge of surrounding road and traffic conditions, which improves decision-making accuracy and motion control effectiveness in ADS \cite{ghorai2022state}. In practice, V2X modules typically enable wireless data exchange with external entities and interface with the ADS software modules to support cooperative driving and coordinated decision-making. In the following, we briefly introduce V2X access technologies, discuss RRM and key communication KPIs, and summarize how V2X enables a shift from ego vehicle-only to cooperative driving paradigms. For detailed technical discussions, the readers are referred to \cite{yurtsever2020survey, zhao2024autonomous}.

\subsubsection{V2X communication technologies}
V2X communication technologies \cite{clancy2024wireless} can be categorized into dedicated short-range communications (DSRC) \cite{kenney2011dedicated} and cellular vehicle-to-everything (C-V2X). 
DSRC is based on IEEE 802.11p and typically provides short one-hop links. Its pure contention-based access enables infrastructure-light deployment but makes performance sensitive to dense traffic: collisions and hidden-node effects may increase end-to-end latency and cause packet delivery rate (PDR) to drop sharply under congestion or at longer distances \cite{gyawali2021challenges}. In contrast, C-V2X (LTE or 5G new radio) can often offer longer one-hop range and higher data rate under comparable scenarios \cite{zhang2022design}. It improves reliability under high channel load by supporting both network-scheduled resource allocation (when cellular infrastructure is available) and distributed semi-persistent sidelink resource selection (e.g., SB-SPS) \cite{clancy2024wireless}, which helps reduce collisions and stabilize latency and PDR degradation under congestion. C-V2X can also be more robust under high mobility due to physical-layer enhancements and NR’s flexible numerologies, at the cost of higher protocol complexity and, in network-scheduled operation, dependence on cellular infrastructure \cite{zhang2022design}. Comprehensive reviews of V2X communication technologies can be found in \cite{clancy2024wireless, gyawali2021challenges}. 

\subsubsection{Radio resource management and communication KPIs}
Beyond the choice of DSRC or C-V2X, practical V2X cooperation critically depends on RRM \cite{masmoudi2019survey}, which allocates and controls radio resources under time-varying topology and bandwidth constraints. Its objective is to satisfy communication key performance indicators (KPIs) \cite{torres2020qos} such as end-to-end latency, reliability, and data rate, which directly affect the timeliness and robustness of V2X message and shared perception information delivery for cooperative driving \cite{huang2025vehicle}. 

\subsubsection{Ego vehicle-only versus V2X-enhanced cooperative scenarios}
Unlike traditional ego vehicle-only scenarios, V2X-enhanced cooperative methods leverage collaboration among multiple agents and infrastructure to enhance perception \cite{zha2025heterogeneous, xu2022v2x, cai2023consensus, liu2024v2x}, prediction \cite{wang2020v2vnet, wang2025cmp, chang2023bev}, and planning \cite{wang2024intercoop, yu2025end, li2024v2x}. Ego vehicle-only scenarios are inherently constrained by occlusion \cite{yuan2021robust}, adverse weather conditions \cite{zhang2021deep}, and sparse long-range observations \cite{zhang2021safe}, which may lead to inaccurate perceptions, unreliable predictions, and decisions and unsafe planning outcomes \cite{yang2025robust, xu2022v2x}. In contrast, V2X-enhanced cooperative scenarios integrate observations from multiple viewpoints to gain a more complete and robust understanding of the environment \cite{xu2022v2x}. This shared perception not only mitigates occlusion and incomplete sensing but also enhances downstream prediction and planning performance. For instance, ego vehicle-only scenarios often fail to anticipate vehicles outside their sensing range or hidden behind obstacles, resulting in underestimated traffic density and inaccurate predictions \cite{chang2023bev, wang2025cmp}. 

Table \ref{table:single-multi-agent} summarizes the differences between ego vehicle-only and V2X-enhanced cooperative paradigms. Beyond sensor coverage and information completeness, it highlights distinctions in interaction modeling, robustness, communication requirements, deployment complexity, and typical application scenarios. Ego vehicle-only paradigms mainly rely on local inference for interaction modeling \cite{li2024v2x}, whereas V2X-enhanced cooperative paradigms can explicitly exchange motion states and planned trajectories \cite{chang2023bev, cai2023consensus}, which improves interaction reasoning. In terms of robustness, ego vehicle-only paradigms are more vulnerable to sensor failures and limited visibility, while cooperative systems benefit from aggregating observations across multiple agents. However, these advantages come at the cost of increased communication demands and deployment complexity \cite{wang2024intercoop, cai2023consensus}, as multi-agent cooperation requires reliable, low-latency communication and system-level synchronization. Consequently, ego vehicle-only paradigms are more suitable for structured highways, whereas V2X-enhanced cooperative paradigms are particularly effective in complex urban intersections, merging zones, and occluded environments \cite{zha2025heterogeneous, chang2023bev}.

At the system level, V2X-based cooperative driving also involves multiple interrelated workloads, such as workload scheduling, computation offloading, and edge-cloud-vehicle orchestration. However, these system-level multi-task optimization methods fall outside the scope of this survey. Our focus is on model-level MTL, where multiple learning objectives are jointly optimized through shared deep neural network architectures. In contrast, system-level methods \cite{li2024eptask, ma2024latency, shao2023multi, zhu2024multi, li2025reinforcement} primarily address workload execution and coordination at the orchestration layer, rather than learning multiple objectives with explicit cross-task coupling. Nevertheless, communication-centric studies that explicitly adopt deep MTL architectures for optimizing network KPIs and RRM decisions are included and discussed in Section \ref{sectionivd}.

\section{Multi-task learning}
\label{section3}
This section provides an overview of MTL. We first formulate the MTL problem and categorize architectural paradigms into hard, soft, and hybrid parameter sharing.  Next, we review representative optimization strategies. Finally, we summarize key advantages of applying MTL in ADS and present a taxonomy of deep MTL methods in CAVs, which serves as the organizing framework for Sections \ref{sec:mtl_perception}–\ref{sectionivd}.

\begin{table*}[t]
    \centering
    \caption{Comparison of three parameter sharing paradigms}
    \label{tab:param_comparison_en}
    \begin{tabular}{lcccc}
    \toprule
    Paradigm & Parameters & Inference Speed & Task Conflict Sensitivity & Application Scenario \\
    \midrule
    Hard-parameter sharing   & Low    & Fast  & High   & Homogeneous tasks, resource-limited \\
    Soft-parameter sharing   & High   & Slow  & Low    & Heterogeneous tasks, sufficient compute \\
    Hybrid-parameter sharing & Medium & Medium & Medium & Complex tasks need balancing speed and performance\\
    \bottomrule
    \end{tabular}
\end{table*}

\subsection{Problem Formulation}
Given $n$ tasks, the learning objective for each task is represented as $\{L_i\}_{i=1}^n$, where we assume all subtasks are related. MTL aims to improve performance across all tasks simultaneously by leveraging knowledge contained in all or some of the tasks \cite{zhang2022surveyzhang}. The overall learning objective can be formulated as minimizing the combined weighted loss:
\begin{equation}
\min_{\Theta, \{\theta_i\}_{i=1}^n} \sum_{i=1}^n \alpha_i L_i\left(f_i(X; \Theta, \theta_i), Y_i\right)
\label{formula:MTL}
\end{equation}
where $X$ is a batch of inputs. $Y_i$ is the ground truth for the $i^{th}$ task corresponding to the batch inputs. $L_i$ is a loss function specific to task $i$. $\Theta$ is the shared layer's parameters, which are common across all tasks, while $\theta_i$ is the task-specific layer's parameters for task $i$. $f_i$ is the forward function for task $i$, where $f_i(X)$ produces the predictions $\hat{Y}_i$ for inputs $X$ based on both the shared and task-specific layers. $\alpha_i$ is the weighting factor for the loss of task $i$, reflecting the relative importance of each task in the overall learning process.

To minimize the overall weighted loss, the optimization is performed jointly over the shared parameters $\Theta$ and all task-specific parameters $\{\theta_i\}_{i=1}^{n}$. This process aims to improve the comprehensive performance across tasks while ensuring generalizability and robustness.

\subsection{Architectural Paradigms}
\label{subsecArchitectural}
MTL consists of shared components that capture common representations across tasks and task-specific components that learn representations unique to each task. The main challenge in MTL design is determining how to share knowledge between tasks in a way that maximizes generalization. To address this challenge, there are three key paradigms, including hard-parameter \cite{wu2022yolop, zhan2024yolopx, wang2024you, yang2020multi}, soft-parameter \cite{misra2016cross, pahari2022multi, kodati2025advancing}, and hybrid-parameter \cite{bruggemann2021exploring, lopes2023cross, chen2024umt} sharing. We provide an overview of each paradigm in the following subsections. Additionally, Table \ref{tab:param_comparison_en} provides a summary of the key characteristics of these paradigms.

\begin{figure}[t]
    \centering
    \includegraphics[width=0.6\linewidth]{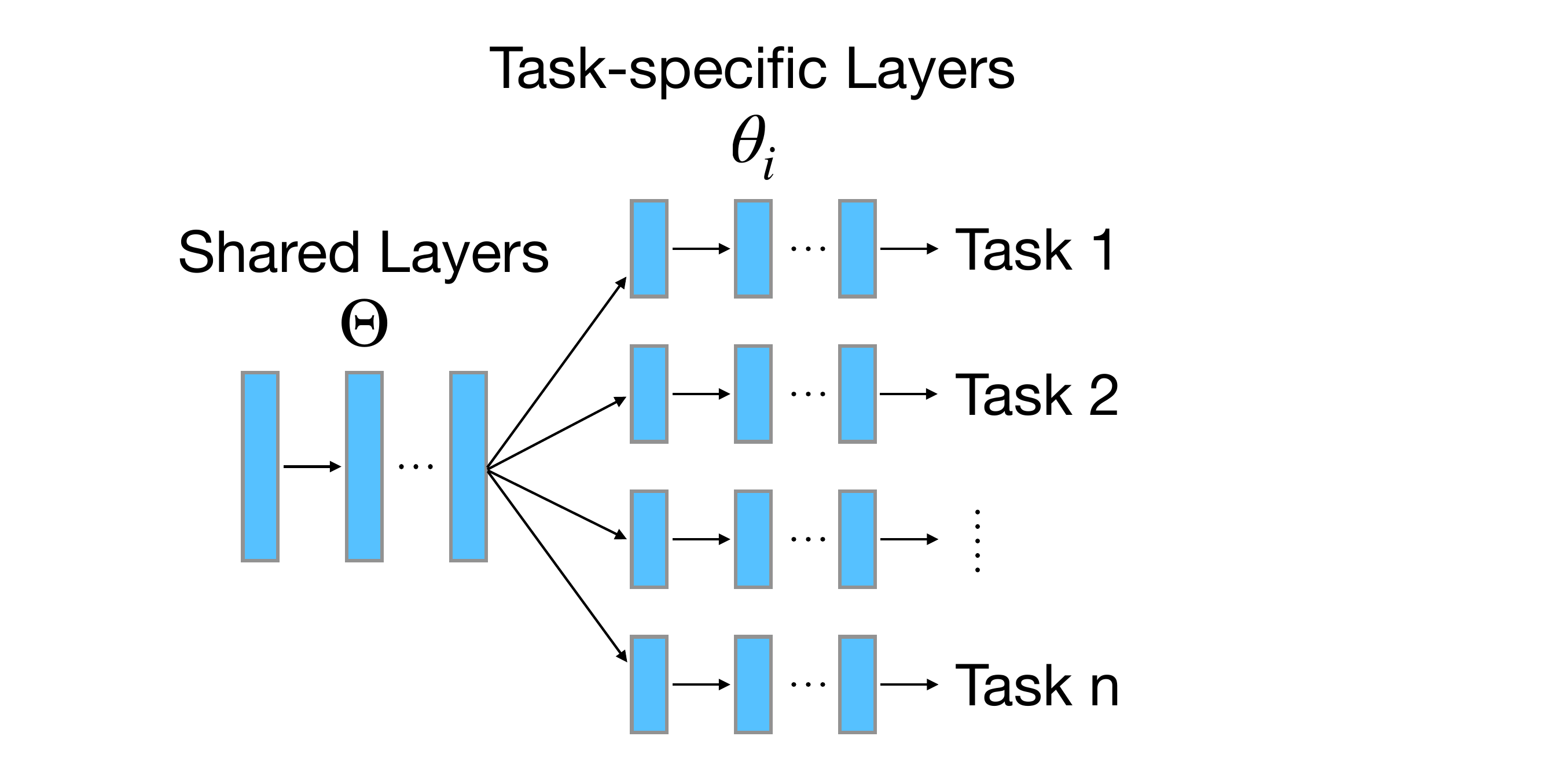}
    \caption{Hard-parameter sharing paradigm.}
    \label{fig:MTL_hard}
\end{figure}

\subsubsection{Hard-parameter sharing} This paradigm is the most commonly used approach in MTL \cite{choi2023dynamic, ruder2017overview}, where the model shares a set of layers, denoted as $\Theta$, across all tasks, and each task has its task-specific layers $\theta_i$ that follow the shared layers (see Fig. \ref{fig:MTL_hard}). Specifically, the shared layers first process the input data, and their output features are then passed to the task-specific layers to produce the final outputs for each task.
This paradigm is computationally effective, as it shares most of the parameters across tasks, which makes it practical for applications with limited computing resources. However, it assumes that all tasks are related, which is not always valid in practice.  
For instance, object detection and depth estimation may exhibit low inter-task correlation. In such cases, hard-parameter sharing can lead to two critical challenges:
\begin{itemize}

\item Negative Transfer: Some or all task performance decreases when irrelevant or conflicting knowledge is transferred across tasks \cite{liu2019loss}.
\item Task Conflict: During training, different tasks may produce conflicting gradients for shared parameters. This makes optimization unstable and may bias one task over others \cite{yu2020gradient}. 
\end{itemize}

\begin{figure}[t]
    \centering
    \includegraphics[width=0.6\linewidth]{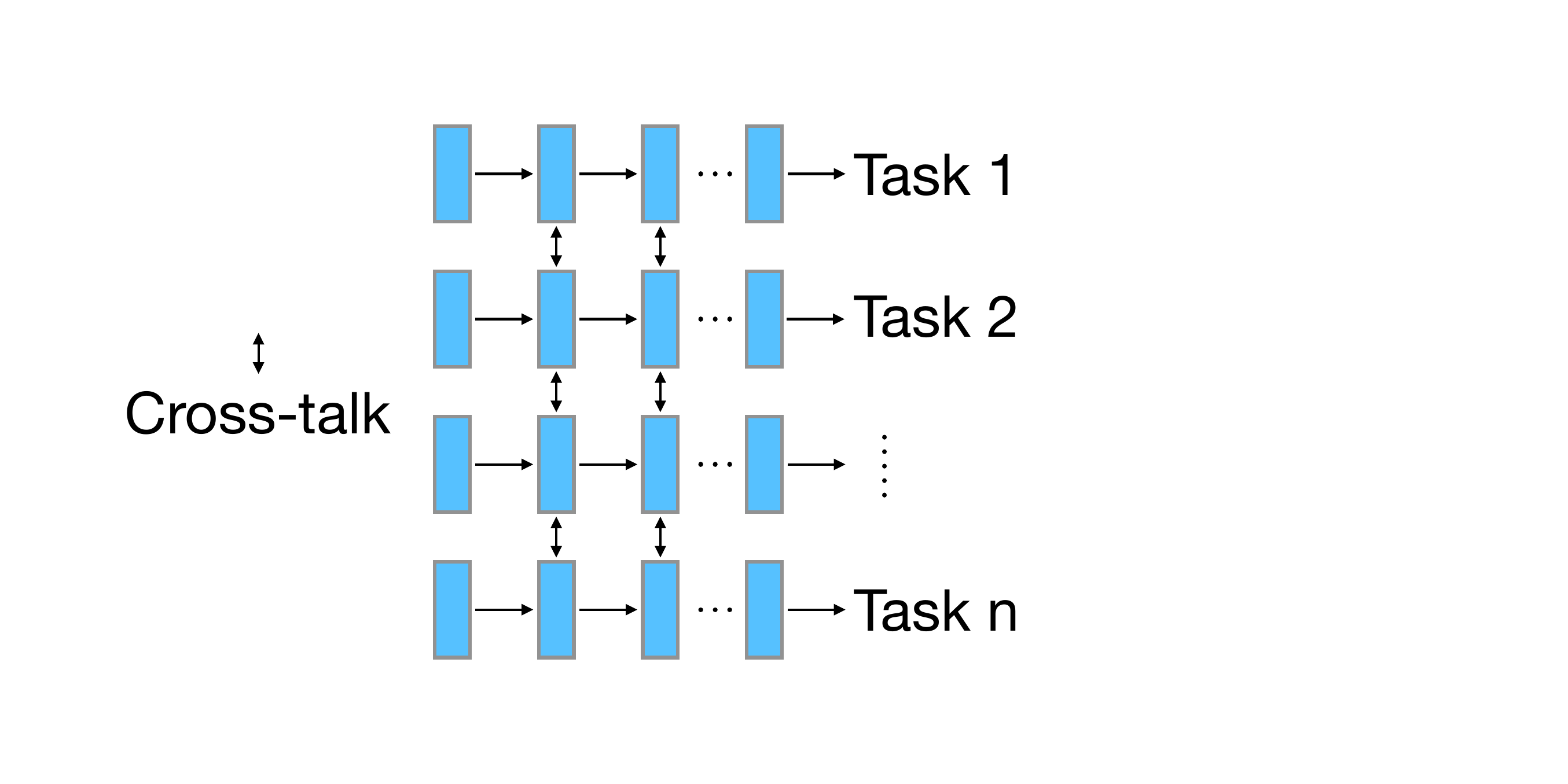}
    \caption{Soft-parameter sharing paradigms.}
    \label{fig:MTL_soft}
\end{figure}
\subsubsection{Soft-parameter sharing} This paradigm (see Fig. \ref{fig:MTL_soft}) uses an alternative approach to the hard-parameter sharing. Each task has its own set of model parameters. Instead of sharing layers, the model exchanges complementary information between different tasks through a mechanism such as cross-talk (also called cross-fusion). For example, the study in \cite{evgeniou2004regularized} uses L2 regularization to constrain the parameters across different tasks. Specifically, an L2 regularization term ($\mathcal{R}_{L2}$) is added into the loss function (Eq. \ref{formula:MTL}) to penalize discrepancies between task-specific weights. L2 regularization penalizes the squared $\ell_{2}$-norm of a parameter vector:
\begin{equation}
\mathcal{R}_{L2} = \|\mathbf{w}\|_{2}^{2} = \sum_{i} w_{i}^{2}.
\end{equation}
In \cite{evgeniou2004regularized}, the L2 penalty is applied not only to individual task parameters but also to the differences between task-specific weights, as follows:
\begin{equation}
\sum_{i=1}^{n} \left\| w_{i} - \frac{1}{n} \sum_{q=1}^{n} w_{q} \right\|^{2},
\end{equation}
where $w_i$ denotes the parameters of the $i$-th task and $\frac{1}{n}\sum_{q=1}^{n} w_{q}$ is the mean parameter vector across tasks, where $q$ is a summation index. This explicitly discourages large discrepancies across tasks and thus enables soft parameter sharing. However, regularization cannot autonomously choose which information to share \cite{li2021empirical}. To address this challenge, Misra et al. \cite{misra2016cross} propose cross-stitch units to dynamically learn linear combinations of task-specific activations. This allows the model to automatically learn the degree of sharing between different tasks. Tian et al. \cite{tian2024unite} propose a plug-and-play module to capture both cross-task consistent and complementary features by computing a cross-task similarity matrix and fusing it with task-specific features via 1$\times$1 convolution. Unlike hard-parameter sharing, soft sharing allows each task flexibility to learn its parameters while maintaining beneficial inter-task communication. This flexibility could effectively alleviate the negative transfer \cite{choi2023dynamic}. While this design provides flexibility, it also introduces scalability concerns. Since each task maintains a full set of parameters, the size of the overall model (i.e., sum of all task-specific model parameters and all the cross-talk modules parameters) tends to grow linearly with the number of tasks \cite{vandenhende2022multi}, which can be problematic when working with limited computational resources.

\begin{figure}[tp]
    \centering
    \includegraphics[width=0.7\linewidth]{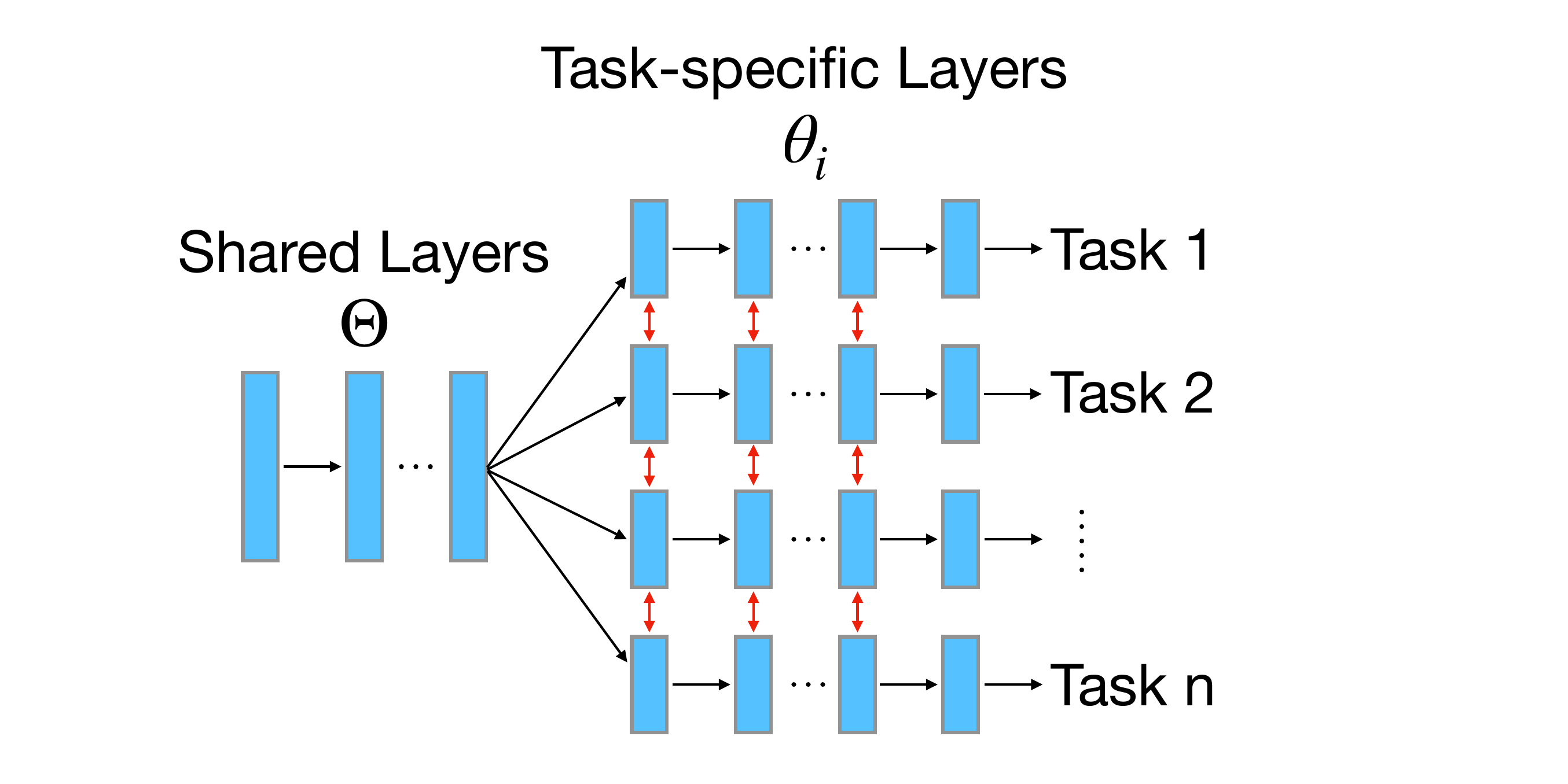}
    \caption{Hybrid-parameter sharing paradigms.}
    \label{fig:MTL_hybrid}
\end{figure}

\subsubsection{Hybrid-parameter sharing} This paradigm (see Fig. \ref{fig:MTL_hybrid}) combines the ideas from both hard and soft-parameter sharing. It consists of a shared backbone that learns common representations from the input, followed by task-specific decoders to refine these features for individual tasks. Additionally, cross-talk mechanisms from soft-parameter sharing are integrated into the task-specific components to enable selective information exchange between task-specific parameters across tasks. 
Several studies have proposed different ways to implement hybrid-parameter sharing. For example, Bruggemann et al. \cite{bruggemann2021exploring} propose the adaptive task-relational context module, which employs attention mechanisms and neural architecture search to automatically optimize cross-task context interactions in multi-task dense prediction. Similarly, Lopes et al. \cite{lopes2023cross} propose a cross-task attention mechanism combining correlation-guided attention and self-attention, fused via learnable channel-wise weights. Their cross-task attention module is integrated into a multi-task exchange block. This block can adaptively refine features across tasks. Additionally, Chen et al. \cite{chen2024umt} introduce an inter-task joint-attention fusion module in the ADS task decoder that dynamically combines features from all tasks' self-attention outputs. This allows cross-task interaction while maintaining parameter efficiency.

\subsection{Optimization Strategies}
Integrating multiple tasks within a single model does not always ensure effective joint learning. One primary challenge in MTL is balancing the optimization dynamics to allow all tasks to progress steadily. Without proper balance, one task may dominate. Other tasks then fail to learn effectively, and the overall performance decreases. To address this issue, several techniques such as loss weighting, gradient-based conflict resolution, and multi-objective optimization have been introduced. 

\subsubsection{Loss weighting}
The objective of MTL involves a weighted combination of task-specific losses (see Eq. \ref{formula:MTL}), where the weighting factor $\alpha_i$ controls the contribution of each task to the overall loss. When one task's loss becomes large, it may overshadow the losses of other tasks. Therefore, it is important to set appropriate weights for each task. A simple yet effective approach is manually tuning the weights through extensive experimentation. However, if the number of tasks is too much, tuning each loss weight becomes difficult. To address this issue, several studies \cite{wu2024adaptive, kendall2018multi, chen2018gradnorm, jha2020adamt, liu2019end} have developed adaptive methods to adjust the weights during training. One popular method is the uncertainty-based weighting mechanism developed by Kendall et al. \cite{kendall2018multi}, where each task's loss is scaled inversely to its homoscedastic uncertainty (a learnable parameter that reflects task-specific noise). Specifically, tasks with high uncertainty are down-weighted. GradNorm \cite{chen2018gradnorm} is another popular method, which obtains the gradient norms from each task’s loss and balances these norms, making sure no task lags significantly. In contrast, Jha et al. \cite{jha2020adamt} normalize the average gradient magnitudes with respect to each task's parameters. This method prioritizes tasks with higher gradients to ensure convex combination stability. Recently, Wu et al. \cite{wu2024adaptive} proposed to adaptively adjust task weights by evaluating each task’s relative inverse training rate, combining loss decay speed and normalized loss magnitude. This prioritizes tasks with slower convergence or higher difficulty, preventing their dominant gradients from overwhelming others.

\subsubsection{Gradient conflict mitigation}
During the training stage, the shared parameters receive the gradient updates information from different tasks, which may point in different or even conflicting directions in parameter space. Such conflicts can cause one task to interfere with another, leading to unstable training or suboptimal convergence. Cosine similarity is commonly used to quantify the alignment between task gradients. It evaluates whether the directions of gradients for two tasks, ($\bm{g_i}$ and $\bm{g_j}$), are compatible. The cosine similarity: 
\begin{equation}
\text{cosine\_similarity} = \cos \theta = \frac{ \bm{g_i} \cdot \bm{g_j} }{ \| \bm{g_i} \| \| \bm{g_j} \| }.
\end{equation}

A low or negative $\cos \theta$ indicates that the gradients are misaligned and present potential interference between the tasks. Identifying and mitigating such conflicts can achieve a stable and balanced multi-task optimization.
Various gradient correction methods have been proposed \cite{yu2020gradient, liu2021conflict, chai2022model, navon2022multi, ji2025variable}. PCGrad \cite{yu2020gradient} projects conflicting gradients onto each other's normal planes to eliminate interfering components. Gradients are altered only when they conflict. Each task’s update does not hurt the others while maintaining constructive interaction. Following the PCGrad, a model-agnostic method \cite{chai2022model} mitigates conflicting gradients in MTL by defining a gradient interfering direction and clipping conflicting gradient components to balance task optimization. Similarly, conflict-averse gradient descent \cite{liu2021conflict} dynamically adjusts the update direction to maximize the worst-case improvement across tasks while constraining updates within a neighborhood of the average gradient. In contrast, Navon et al. \cite{navon2022multi} propose using the Nash bargaining solution to derive a scale-invariant, Pareto-optimal update direction that balances task gradients proportionally. This ensures fairness and convergence guarantees.
However, these methods increase GPU memory consumption because they require computing and storing gradients for each task separately, rather than using a single backward pass, reducing the maximum batch size and increasing the required training time.

\subsubsection{Multi-objective optimization} 
This category of optimization considers MTL as a multi-objective optimization problem, where each task loss is an objective, and there is usually no single solution that is best for all objectives unless they are perfectly related. Sener and Koltun \cite{sener2018multi} formulate MTL as a multi-objective optimization problem to seek Pareto optimal solutions. The aim is to minimize a vector-valued loss:
\begin{equation}
\min_{\boldsymbol{\Theta}, \boldsymbol{\theta}_1, \ldots, \boldsymbol{\theta}_n} \mathbf{L} = \left( \hat{\mathcal{L}}_1(\boldsymbol{\Theta}, \boldsymbol{\theta}_1), \ldots, \hat{\mathcal{L}}_n(\boldsymbol{\Theta}, \boldsymbol{\theta}_n) \right)^\intercal,
\end{equation}
where $\Theta$ is shared parameters by all tasks, $\theta_1, ... ,\theta_n$ are task-specific parameters for tasks $1, ..., n$. To solve this, they adapt the multiple gradient descent algorithm (MGDA), which computes coefficients $\{\alpha_t\}$ by solving:
{\small
\begin{equation}
\min_{\alpha_{1}, \ldots, \alpha_{n}} \left\{ \left\| \sum_{t=1}^{n} \alpha_{t} \nabla_{\Theta} \hat{L}_{t}(\Theta, \theta_{t}) \right\|_{2}^{2} \;\middle|\; \sum_{t=1}^{n} \alpha_{t} = 1, \; \alpha_{t} \geq 0 \; \forall t \right\}.
\label{formula:MGDA}
\end{equation}
}
To avoid $n$ backward passes in MGDA, they introduce MGDA-UB, which optimizes an upper bound via shared representations $\mathbf{Z} = g(\mathbf{X}; \boldsymbol{\Theta})$. The resulting optimization problem is:
{\small
\begin{equation}
\min_{\alpha_{1}, \ldots, \alpha_{n}} \left\{ \left\| \sum_{t=1}^{n} \alpha_{t} \nabla_{\mathcal{Z}} \hat{L}_{t}(\Theta, \theta_{t}) \right\|_{2}^{2} \;\middle|\; \sum_{t=1}^{n} \alpha_{t} = 1, \; \alpha_{t} \geq 0 \; \forall t \right\}.
\end{equation}
}
Under full-rank assumptions of $\frac{\partial \mathbf{Z}}{\partial \boldsymbol{\Theta}}$, MGDA-UB guarantees Pareto optimality. However, they find a single Pareto solution for MTL. Lin et al. \cite{lin2019pareto} extend Sener and Koltun's work by decomposing the problem into preference-guided subproblems to enable the generation of diverse Pareto-optimal solutions that represent distinct trade-offs across tasks. Following previous work, Ma et al. \cite{ma2020efficient} advance MTL by proposing continuous Pareto exploration, which constructs locally smooth Pareto sets through second-order analysis and Krylov subspace methods. This approach generates dense Pareto fronts that capture a wider range of trade-offs between conflicting tasks while scaling effectively to large-scale neural networks. Additionally, Momma et al. \cite{momma2022multi} integrates user preferences with Pareto stationarity, proposing the extended weighted Chebyshev method to efficiently discover Pareto optimal solutions aligned with preferences or reference models to reduce exploration costs from $\Omega(m)$ to $O(1)$ while achieving competitive performance.

\subsection{Motivations for Applying MTL in CAVs}

CAVs are required to perform multiple tasks simultaneously while operating under strict onboard hardware constraints. Beyond onboard-only ADS, CAVs can use V2X connectivity to incorporate information exchanged across vehicles and infrastructure, which introduces system-level constraints such as communication latency, reliability, and bandwidth limits. MTL addresses these challenges by joint learning of multiple tasks within a unified model, where inputs can be from onboard sensing or multi-agent cooperative information when available. This shared learning reduces resource consumption, promotes knowledge transfer, and streamlines the integration of new tasks. In this subsection, we highlight three key advantages of applying MTL in CAVs: improved computational efficiency, enhanced task interaction, and greater flexibility in model updates. We further present our proposed taxonomy of deep MTL methods in CAVs.

\begin{table}[t]
    \centering
    \caption{Proposed taxonomy for MTL methods in CAVs by functional domain and operational context.}
    \label{tab:mtl_taxonomy}
    \includegraphics[width=0.90\linewidth]{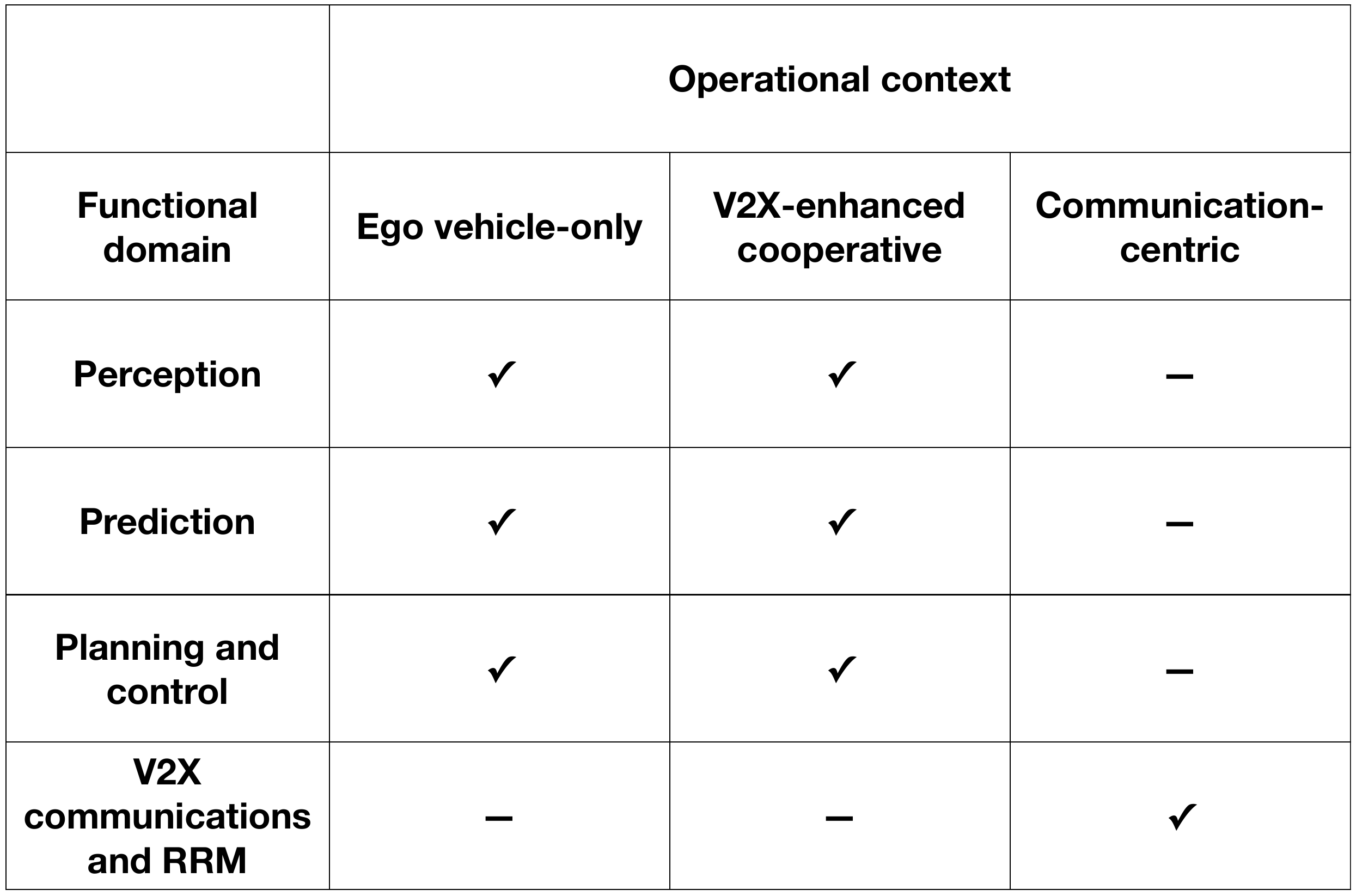}
\end{table}

\subsubsection{Computational efficiency and resource optimization}
MTL reduces computational costs, memory usage, and energy consumption through sharing parameters across tasks \cite{wang2024you, miraliev2024real, guo2023research}. This efficiency is critical for resource-constrained edge devices and enables real-time processing, which requires frames per second (FPS) over 30 \cite{wang2023centernet, bavirisetti2023multi} while maintaining accuracy. For instance, Iliadis et al. \cite{iliadissignal} show that a unified MTL model avoids maintaining $T$ separate single-task models and reduces the computational complexity from $\mathcal{O}(T \cdot L \cdot H \cdot D)$ to $\mathcal{O}(L \cdot H \cdot D)$ by reusing shared representations. In V2X-enhanced cooperation, shared representations can improve communication efficiency and reduce redundancies in multi-agent feature fusion \cite{yan2025multi}. For example, Eldeeb et al. \cite{eldeeb2024multi} report up to about 89\% reduction in transmitted bytes via a shared semantic encoder while maintaining comparable or improved reconstruction and classification performance. Additionally, MTL can further save the training cost by combining with transfer learning \cite{wang2023sparse}. Specifically, pre-train the model on a large-scale dataset (such as ImageNet \cite{deng2009imagenet}) to learn a general representation and fine-tune it on downstream tasks. This is especially useful for Transformer-based models, which typically require a longer training time compared to CNN-based models.

\subsubsection{Task synergy and knowledge transfer}
MTL exploits the implicit synergy between tasks to enhance performance \cite{liang2022effective, wu2022yolop}. For example, semantic segmentation masks improve object detection by providing contextual boundaries, while object detection outputs guide lane segmentation through spatial constraints \cite{zhang2021loss}. This cross-task knowledge transfer mitigates the need for exhaustive labeled datasets for individual tasks. It combines the labeled data from all tasks, effectively serving as a form of data augmentation, to build a more accurate model for each task. Additionally, training with multiple tasks could regularize the model, further reducing overfitting for each task \cite{kim2020multi, miraliev2024real, yang2024efficient}.

\subsubsection{Scalability and modularity}

MTL offers a modular and extensible architecture design that is particularly beneficial for CAVs, where perception or sensor requirements constantly evolve. By decoupling a model into a shared backbone and multiple task-specific heads, it can flexibly incorporate new tasks \cite{ishihara2021multi, rao2024camera, wang2024you}. Additionally, in V2X-enhanced cooperative scenarios, vehicles and infrastructure units can adopt different sensor configurations \cite{xiang2023multi, zhao2024coopre}. The modular design facilitates the integration of new sensors as additional inputs without modifying the entire network \cite{ye2023fusionad, rao2024enhancing}. This modularity also simplifies maintenance and enables targeted fine-tuning for individual tasks as requirements evolve.

\begin{table*}[t]
\centering
\caption{Overview of ego vehicle-only MTL for perception works.}
\label{tab:onboard_perception_overview_all}
\footnotesize
\setlength{\tabcolsep}{2.5pt}
\renewcommand{\arraystretch}{1.12}

\resizebox{\textwidth}{!}{
\begin{tabular}{l c c c P{0.16\textwidth} ccccccccc}
\toprule
Ref. & Year & Method & Share & Dataset(s) &
Det & SemSeg & InsSeg & Lane & Drivable & Depth & Pose & Dist & COE \\
\midrule
\cite{petrovai2019multi} & 2019 & R-CNN & H & Cityscapes &  & \checkmark & \checkmark &  &  &  &  &  &  \\
\cite{yang2020multi} & 2020 & R-CNN & H & KITTI & \checkmark &  &  &  &  &  &  &  &  \\
\cite{rinchen2023scalable} & 2023 & R-CNN & H & BDD100K & \checkmark &  &  &  &  &  &  &  &  \\
\cite{fang2023improved} & 2023 & R-CNN & H & Cityscapes & \checkmark &  & \checkmark &  &  &  &  &  &  \\
\cite{teichmann2018multinet} & 2018 & C-CNN & H & KITTI & \checkmark & \checkmark &  &  &  &  &  &  &  \\
\cite{leang2020dynamic} & 2020 & C-CNN & H & KITTI; Cityscapes; WoodScape & \checkmark & \checkmark &  &  &  &  &  &  &  \\
\cite{zhao2023drmnet} & 2023 & C-CNN & H & BDD100K & \checkmark &  &  & \checkmark & \checkmark &  &  &  &  \\
\cite{miraliev2024real} & 2024 & C-CNN & H & BDD100K & \checkmark &  &  & \checkmark & \checkmark &  &  &  &  \\
\cite{chen2024umt} & 2024 & C-CNN & Hy & Cityscapes; NYUv2; BDD100K & \checkmark & \checkmark &  & \checkmark & \checkmark & \checkmark &  &  &  \\
\cite{wu2022yolop} & 2022 & YOLO & H & BDD100K & \checkmark &  &  & \checkmark & \checkmark &  &  &  &  \\
\cite{han2022yolopv2} & 2022 & YOLO & H & BDD100K & \checkmark &  &  & \checkmark & \checkmark &  &  &  &  \\
\cite{lee2022joint} & 2022 & YOLO & H & BDD100K & \checkmark &  &  & \checkmark & \checkmark &  &  &  &  \\
\cite{guo2023research} & 2023 & YOLO & H & BDD100K & \checkmark &  &  & \checkmark & \checkmark &  &  &  &  \\
\cite{zhang2024parallel} & 2024 & YOLO & H & BDD100K; KITTI & \checkmark &  &  & \checkmark & \checkmark &  &  &  &  \\
\cite{zhan2024yolopx} & 2024 & YOLO & H & BDD100K & \checkmark &  &  & \checkmark & \checkmark &  &  &  &  \\
\cite{fang2024yolomh} & 2024 & YOLO & H & BDD100K & \checkmark &  &  & \checkmark & \checkmark &  &  &  &  \\
\cite{wang2024you} & 2024 & YOLO & H & BDD100K & \checkmark &  &  & \checkmark & \checkmark &  &  &  &  \\
\cite{li2025mtpnet} & 2025 & YOLO & H & BDD100K & \checkmark &  &  & \checkmark & \checkmark &  &  &  &  \\
\cite{chen2018multi} & 2018 & SSD & H & KITTI & \checkmark &  &  &  &  &  &  & \checkmark &  \\
\cite{kumar2021vru} & 2021 & SSD & H & TDUP & \checkmark &  &  &  &  &  & \checkmark &  &  \\
\cite{liu2021mtnas} & 2021 & SSD & H & KITTI; Cityscapes; (Waymo\,+\,BDD100K); (Cityscapes\,+\,BDD100K) & \checkmark & \checkmark &  &  &  &  &  &  &  \\
\cite{mohamed2021spatio} & 2021 & HC-Trans & Hy & KITTI MOD & \checkmark & \checkmark &  &  &  &  &  &  &  \\
\cite{wang2023sparse} & 2023 & HC-Trans & Hy & BDD100K & \checkmark &  &  & \checkmark & \checkmark &  &  &  &  \\
\cite{li2024cutransnet} & 2024 & HC-Trans & H & BDD100K & \checkmark &  &  & \checkmark & \checkmark &  &  &  &  \\
\cite{wang2025rmt} & 2025 & HC-Trans & H & BDD100K & \checkmark &  &  & \checkmark & \checkmark &  &  &  &  \\
\cite{bavirisetti2023multi} & 2023 & Trans & H & KITTI &  & \checkmark &  &  &  & \checkmark &  &  &  \\
\cite{chen2025m3net} & 2025 & Trans & H & (nuScenes + OpenOccupancy) & \checkmark & \checkmark &  &  &  &  &  &  & \checkmark \\
\cite{liang2022effective} & 2022 & VLM & H & BDD100K & \checkmark & \checkmark &  & \checkmark & \checkmark &  &  &  &  \\
\cite{liang2023visual} & 2023 & VLM & H & BDD100K & \checkmark & \checkmark &  & \checkmark & \checkmark &  &  &  &  \\
\bottomrule
\end{tabular}}

\vspace{2pt}
\footnotesize
\raggedright
\justifying
\textit{Note:} \textbf{Det}: Object detection. \textbf{SemSeg}: Semantic segmentation. \textbf{InsSeg}: Instance segmentation. \textbf{Lane}: Lane line segmentation. \textbf{Drivable}: Drivable area segmentation. \textbf{Depth}: Depth estimation. \textbf{Pose}: Pose estimation. \textbf{Dist}: Object-level distance classification (e.g., near/mid/far). \textbf{COE}: Current occupancy estimation. \textbf{H}: Hard-parameter sharing. \textbf{Hy}: Hybrid-parameter sharing. \textbf{C-CNN}: Conventional CNN. \textbf{HC-Trans}: Hybrid CNN-Transformer architecture.
\end{table*}

\subsubsection{Taxonomy of Deep MTL in CAVs}
In CAVs, MTL is applied not only to ego vehicle ADS software layers, but also to learning problems arising within the V2X communication stack and network-side RRM. Existing deep MTL methods could be categorized according to their functional domains within CAV systems into four categories: (i) perception, (ii) prediction, (iii) planning and control, and (iv) V2X communications and RRM. The first three categories correspond to the conventional software layers of the ADS. In contrast, the last category focuses on communication-centric MTL problems, where learning objectives are defined within the V2X stack and network control layer. Based on these considerations, we propose a taxonomy of MTL methods in CAVs by functional domain and operational context (see Table \ref{tab:mtl_taxonomy}).

For the perception, prediction, and planning and control domains, the reviewed MTL methods are discussed under two operational contexts: ego vehicle-only and V2X-enhanced cooperative. Ego vehicle-only MTL assumes that all tasks are executed locally using only onboard sensors and computing platforms. In contrast, V2X-enhanced cooperative MTL leverages information exchanged through V2V, V2I, or V2N links to support cooperative perception, prediction, planning, and control among multiple agents. This distinction is particularly important in CAVs, where the same learning task can be addressed either as a purely onboard problem or as a multi-agent cooperative problem, leading to different design constraints and architectural choices. In practice, V2X-enhanced cooperation is not always available \cite{hobert2015enhancements} or reliable \cite{ren2024interruption}. Therefore, it is necessary to discuss ego vehicle-only operation in CAVs. Although planning and control are conventionally treated as separate ADS modules, they are considered jointly in our proposed taxonomy due to the relatively limited number of deep MTL studies targeting these functions, while planning-centric and control-centric approaches are still distinguished in the corresponding subsections for clarity. The V2X communications and RRM domain are discussed as a class of communication-centric MTL problems, rather than under ego vehicle-only and V2X-enhanced cooperative paradigms, since its learning targets are inherently defined within V2X links and network control loops.

All the works in Sections \ref{sec:mtl_perception} --\ref{sec:mtl_planning_control} implement deep MTL at the model level, where multiple ADS functions are coupled within a single neural architecture, rather than being trained as completely separate networks. In addition, Section~\ref{sectionivd} reviews communication-centric MTL, where the multi-task coupling is realized through a unified learning model or policy within the V2X communication and resource-control stack, and the learned representations or policies are primarily evaluated using communication KPIs and resource-control outcomes.

Existing MTL studies are heavily skewed towards perception tasks, while prediction, planning, control, and V2X communications and RRM remain relatively under-explored. This imbalance is evident throughout the following Sections \ref{sec:mtl_perception}--\ref{sectionivd}. Across these tasks, V2X-enhanced cooperative MTL is still at an early stage, with a smaller body of literature compared to ego vehicle-only MTL. 
One contributing factor is that, although DL has driven major advances in ADS since the early 2010s \cite{grigorescu2020survey}, reliable V2X communication infrastructure and large-scale cooperative driving testbeds are only recently becoming available and remain under active deployment \cite{yang2022autonomous}. In addition, V2X-enhanced MTL introduces extra degrees of complexity, such as multi-agent synchronization, communication reliability, cross-vehicle calibration, and jurisdiction-dependent safety, which collectively slow the development cycle relative to ego vehicle-only MTL.

\section{MTL for Perception}
\label{sec:mtl_perception}
Perception-oriented MTL in CAVs aims to use a single model to perform multiple perception tasks under strict real-time constraints in either the ego vehicle-only paradigm (when V2X is unavailable) or the V2X-enhanced cooperative paradigm. These tasks include object detection, semantic and instance segmentation, drivable area segmentation, depth or occupancy prediction, and so on. The main challenge is sharing backbone networks and intermediate features across tasks while mitigating negative transfer and satisfying latency requirements. In cooperative paradigms, additional challenges arise from occlusions, localization and pose misalignment among agents, and unreliable V2X links. In the following subsections, the first examines ego vehicle-only perception architectures that operate entirely on onboard sensing, while the second focuses on V2X-enhanced cooperative perception, where multi-agent feature fusion and communication constraints play a critical role. Finally, we summarize key lessons learned.

\subsection{Ego Vehicle-Only Perception} 
Ego vehicle-only perception can be broadly grouped from a model architecture viewpoint into CNN-based, Transformer-based, and vision language model (VLM)-based methods. Table \ref{tab:onboard_perception_overview_all} summarizes representative ego vehicle-only perception MTL methods.

\subsubsection{CNN-based methods}
\begin{figure}[t]
    \centering
    \includegraphics[width=\linewidth]{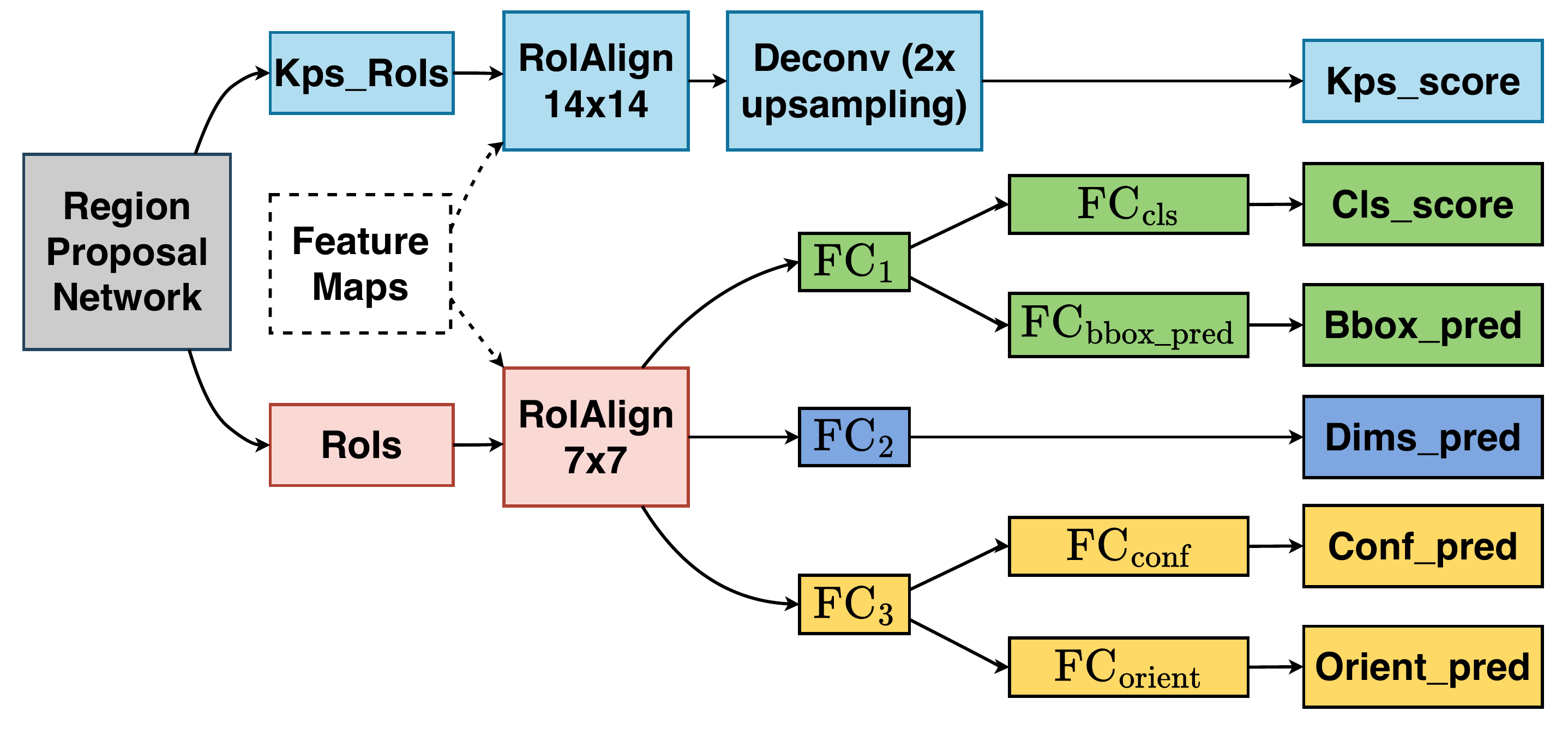}
    \caption{Representative R-CNN family MTL architecture (MT-Faster R-CNN) (adapted from \cite{yang2020multi}.)}
    \label{fig:MT-Faster}
\end{figure}

CNN-based ego vehicle-only perception MTL models follow two main patterns: \textit{(i)} two-stage region-based detectors with multi-task heads, and \textit{(ii)} one-stage detection-centric networks augmented with auxiliary task heads. Two-stage methods are based on Faster R-CNN or Mask R-CNN \cite{ren2016faster, he2017mask}, where a shared backbone and region proposal network extract generic features and region proposals, while multiple region of interest (RoI) heads specialize in different tasks. MT-Faster R-CNN \cite{yang2020multi} is an example of CNN-based two-stage MTL that simultaneously performs 2D and 3D detection, orientation estimation, and key-point detection for monocular driving scenes. The model (see Fig. \ref{fig:MT-Faster}) includes a region of interest alignment layer (RoIAlign) to accurately extract features while preserving spatial alignment. The region proposal network generates two branches of RoI: one predicts classification, dimensions, confidence, and rotation angle, while the other outputs key point scores for 3D detection and orientation estimation. Other related works extend Mask R-CNN to jointly implement instance, semantic, and panoptic segmentation \cite{petrovai2019multi}, or to scale to identify multiple object categories via task-specific RoI heads and multiple region proposal networks, such as traffic lights, signs, pedestrians, and vehicles \cite{rinchen2023scalable}. Fang et al. \cite{fang2023improved} further refine Mask R-CNN for complex traffic scenes by upgrading the backbone to ResNeXt and incorporating improved IoU losses \cite{zheng2022enhancing}. These works illustrate the flexibility of region-based two-stage MTL, but their multi-stage pipelines and high computational costs often limit real-time deployment \cite{carranza2020performance}.

On the other hand, one-stage methods directly regress classes and bounding boxes from dense feature maps to avoid the proposal generation stage and significantly reduce computational cost. This design enables shorter inference time than two-stage methods \cite{carranza2020performance}, making one-stage methods suitable for real-time ego vehicle-only perception in CAVs. MultiNet \cite{teichmann2018multinet} adopts a shared encoder with separate decoders for classification, detection, and semantic segmentation. While Leang et al. \cite{leang2020dynamic} introduce adaptive loss weighting to balance the learning pace among tasks and stabilize multi-task optimization. Recent studies leverage lightweight backbones \cite{miraliev2024real} (e.g., RegNetY \cite{radosavovic2020designing} and MobileNetV3 \cite{howard2019searching}) and multi-scale feature fusion \cite{chen2024umt, zhao2023drmnet}. For instance, UMT-Net \cite{chen2024umt} combines a shared encoder with task-specific self-attention encoders and a joint-attention fusion module, representing a typical hybrid-parameter sharing design that encourages cross-task interaction without fully disentangling task heads.

Within the one-stage family, You Only Look Once \cite{redmon2016you} (YOLO)-based models have become particularly prominent for ego vehicle-only MTL due to their favorable trade-off between inference speed and accuracy. One early work is YOLOP \cite{wu2022yolop} that adopts a hard-parameter sharing with CSPDarknet backbone, spatial pyramid pooling \cite{he2015spatial} and feature pyramid network \cite{lin2017feature} neck, and adds three heads for object detection, drivable area segmentation, and lane line segmentation. YOLOP achieves state-of-the-art on the three tasks of the BDD100K dataset and achieves real-time performance at that time. The following works, such as YOLOPX \cite{zhan2024yolopx} and YOLOPv2 \cite{han2022yolopv2}, refine the backbone, neck, and loss design to improve accuracy and efficiency. A series of works \cite{wang2024you, fang2024yolomh, zhang2024parallel, guo2023research, zhao2023drmnet, lee2022joint, li2025mtpnet} follow YOLOP's hard-parameter sharing architecture, differing mainly in backbone choice, neck structure, and task-balancing strategy. All of these works aim to jointly perform multiple segmentation and detection tasks under real-time constraints.

\begin{figure}[t]
    \centering
    \includegraphics[width=\linewidth]{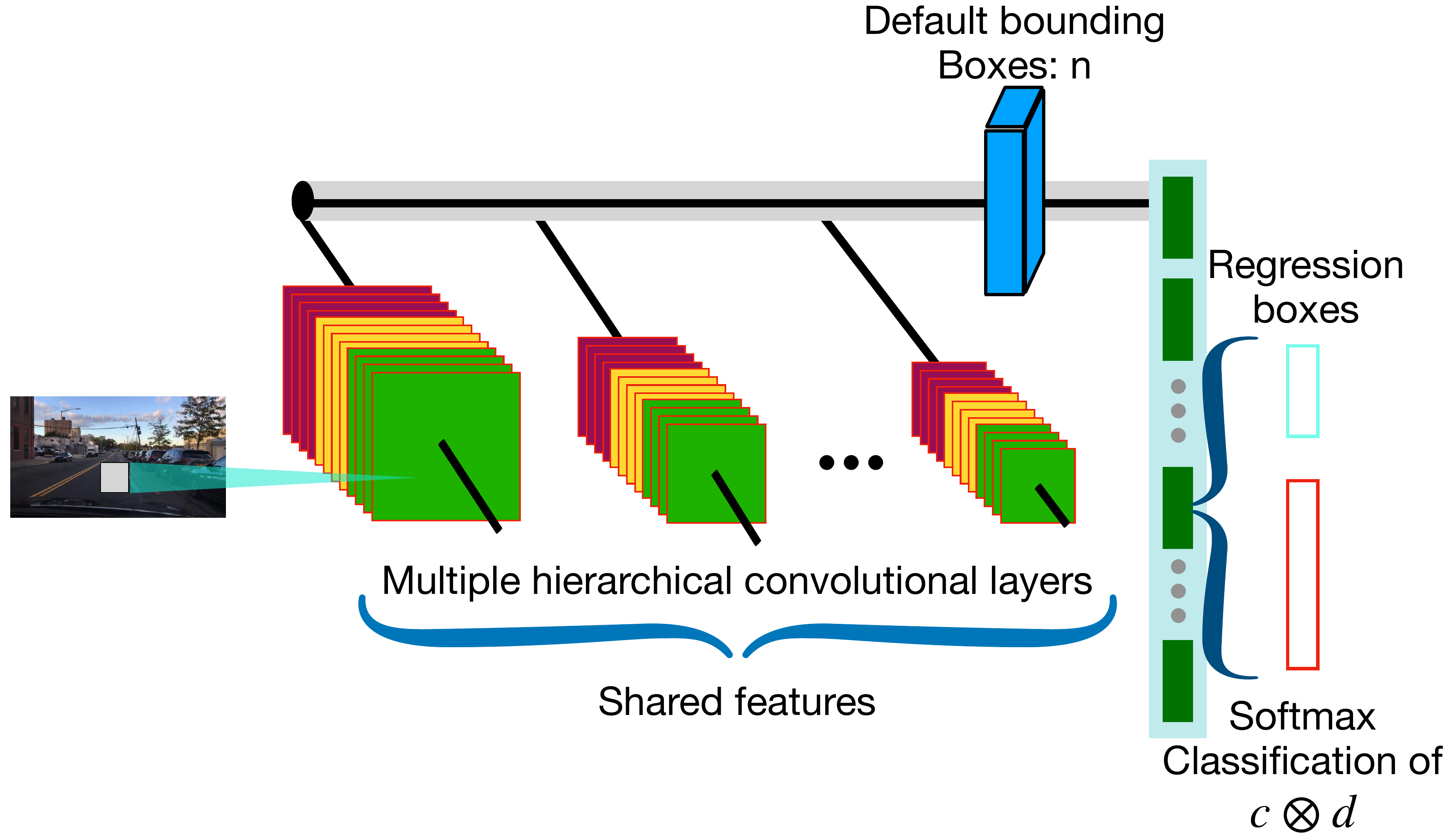}
    \caption{Representative SSD family MTL architecture (CP-MTL SSD) (adapted from \cite{chen2018multi}).}
    \label{fig:CP-MTL}
\end{figure}

Single-shot multi-box detector (SSD) \cite{liu2016ssd} is another one-stage method for multi-task perception. These models typically attach auxiliary tasks, such as distance prediction or pose estimation, to the detection head. For example, CP-MTL SSD \cite{chen2018multi} jointly performs dangerous object detection and distance categorization. The model structure is illustrated in Fig. \ref{fig:CP-MTL}, where output $d$ is the category of an object distance, and $c$ is the object categories. While other works incorporate pose estimation for vulnerable road users \cite{kumar2021vru} or employ neural architecture search to optimize both task-specific branches and the shared multi-task backbone \cite{liu2021mtnas}.
However, SSD has notable limitations in detecting small objects \cite{botezatu2024review, zhai2020df}. As a result, SSD-based methods are less appealing in ADS scenarios, where onboard cameras frequently capture distant objects that appear smaller due to perspective effects.

\subsubsection{Transformer-based methods}
Transformer-based ego vehicle-only MTL methods introduce self-attention mechanisms to better capture long-range dependencies in sequential data \cite{vaswani2017attention}. It allows parallelization during training and overcomes the limitations of traditional recurrent architectures in handling long-term dependencies. 
Bavirisetti et al. \cite{bavirisetti2023multi} adopt a SegFormer-style Transformer encoder shared by semantic segmentation and monocular depth estimation, with very lightweight task decoders, thereby concentrating most capacity in a global vision Transformer backbone \cite{dosovitskiy2020image}. In contrast, M3Net \cite{chen2025m3net} builds multimodal bird's-eye-view (BEV)-centric multi-task perception that jointly performs 3D object detection, BEV map segmentation, and current-frame 3D semantic occupancy estimation from fused camera and LiDAR inputs. M3Net employs a shared Transformer-based fusion module together with task-oriented channel scaling to mitigate gradient conflicts across task heads. Overall, these works demonstrate that pure Transformer designs for perception MTL can provide strong global reasoning and flexible multi-head outputs, but they also incur non-trivial computational cost and require explicit mechanisms to alleviate gradient conflicts during joint optimization \cite{bavirisetti2023multi, chen2025m3net}.

Unlike CNNs, Transformers lack certain inductive biases, such as translation equivariance, spatial invariance, and locality \cite{han2022survey}. As a result, when trained with insufficient data, Transformers may underperform compared to CNNs and exhibit weaker generalization \cite{han2022survey}. However, hybrid models that combine CNNs with Transformers help mitigate these limitations and achieve competitive performance. In particular, hybrid models offer higher throughput while maintaining parameter counts and computational complexity (FLOPs) that are intermediate between those of pure Transformer and pure CNN architectures \cite{han2022survey}. 

\begin{figure}[t]
    \centering
    \includegraphics[width=\linewidth]{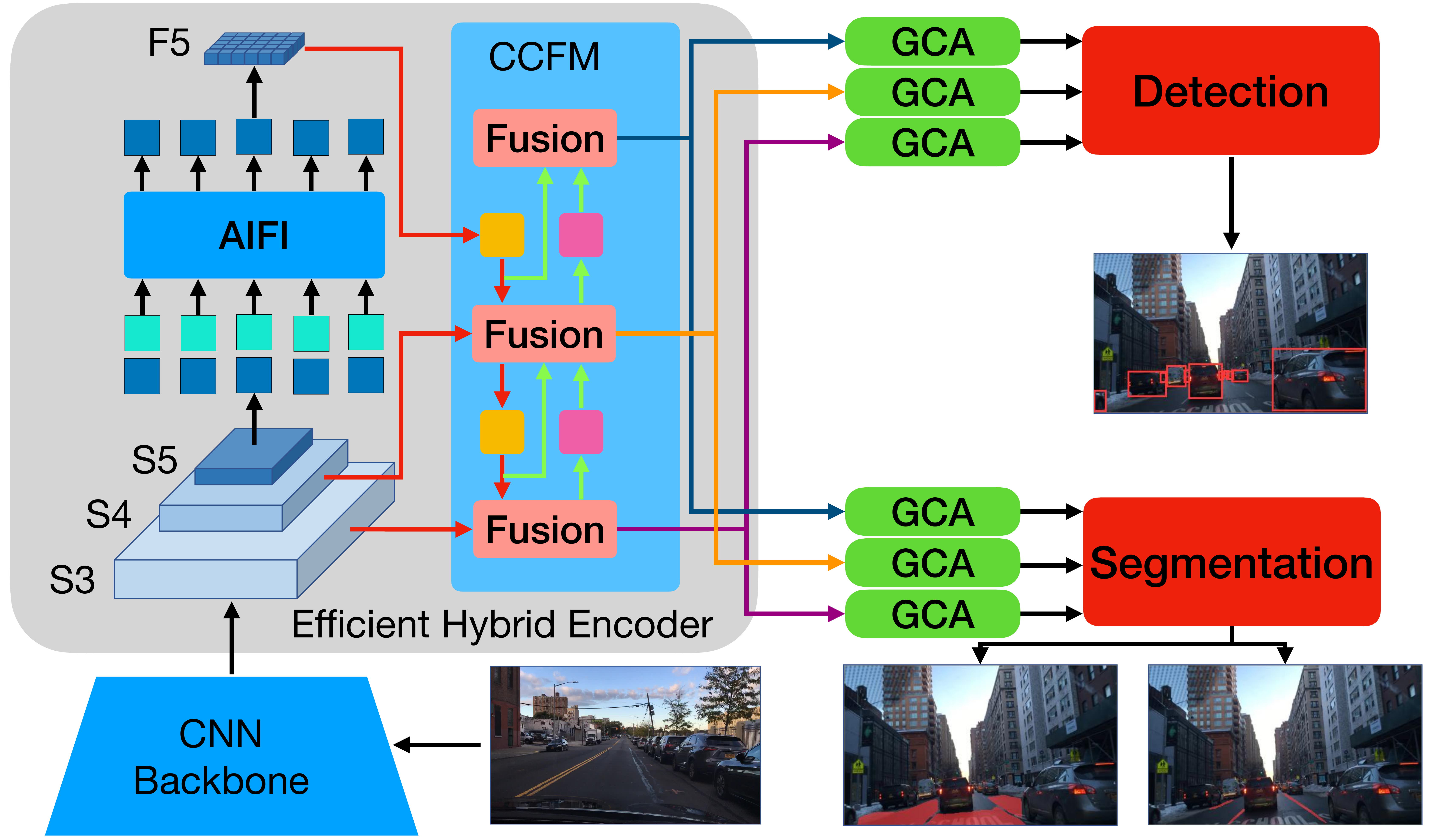}
    \caption{Representative hybrid CNN-Transformer MTL architecture (RMT-PPAD) (adapted from \cite{wang2025rmt}).}
    \label{fig:RMTPPAD}
\end{figure}

Hybrid CNN–Transformer MTL models combine the inductive biases and computational efficiency of CNNs with the global reasoning capability of Transformers to improve multi-task perception accuracy in ego vehicle-only paradigms, while meeting onboard runtime constraints \cite{wang2023sparse, mohamed2021spatio, li2024cutransnet, wang2025rmt}. These models typically employ convolutional encoders for efficient local feature extraction and integrate Transformer blocks to capture global context, temporal dependencies, and inter-task interactions. For instance, a simple and efficient MTL driving perception model, named sparse U-PDP (unified panoptic driving perception) \cite{wang2023sparse}, integrates vehicle detection, lane detection, and drivable-area segmentation using a shared encoder and a unified Transformer-based decoder to model inter-task relationships. It adopts task-aware dynamic convolution kernels and a dynamic interaction module within the Transformer decoder, enabling task-specific feature sampling and richer information exchange across tasks. Similarly, Mohamed et al. \cite{mohamed2021spatio} build on convolutional feature extractors but add a spatio-temporal Transformer that jointly reasons about moving-object detection and segmentation over video sequences. CUTransNet \cite{li2024cutransnet} adopts a U-shaped hybrid encoder in which CNN feature pyramids are strengthened by Transformer blocks before branching into detection and semantic segmentation heads for traffic objects, lane markings, and drivable areas. RMT-PPAD \cite{wang2025rmt} is a hybrid CNN-Transformer MTL design (Fig. \ref{fig:RMTPPAD}) that builds on an RT-DETR-style efficient hybrid encoder and introduces lightweight gate control with an adapter module to mitigate negative transfer across detection and segmentation tasks.

\subsubsection{VLM-based methods}
VLMs augment ego vehicle-only perception MTL by incorporating textual information and language priors into visual representations. Typically, an image encoder $f_{\theta}$ and a text encoder $f_{\phi}$ are pre-trained on large-scale datasets $\mathcal{D} = \{x_n^I, x_n^T\}_{n=1}^N$ and then adapted to downstream driving tasks through prompt learning or lightweight adapters \cite{zhang2024vision}.

\begin{figure}[t]
    \centering  
    \includegraphics[width=\linewidth]{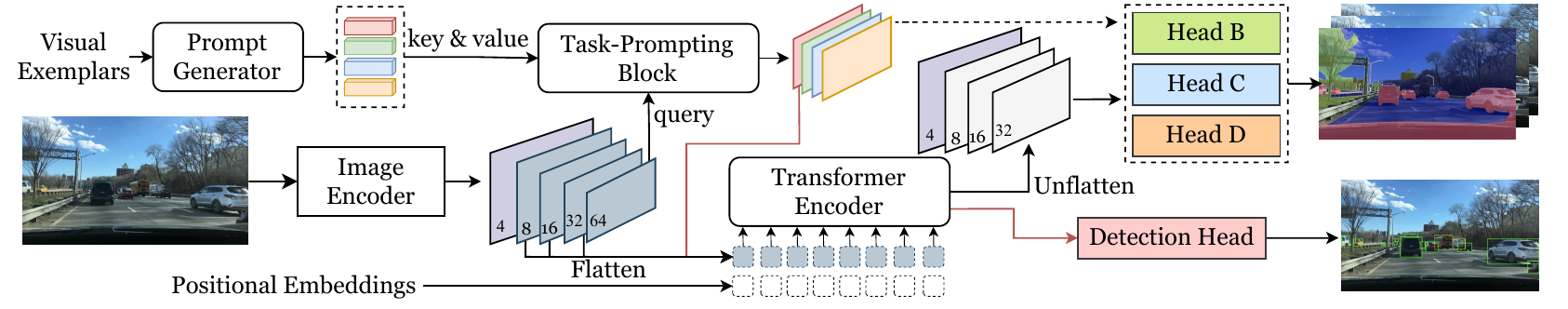}
    \caption{Representative VLM-based MTL architecture (VE-Prompt) (adapted from \cite{liang2023visual}).}
    \label{fig:VE-Prompt}
\end{figure}

Liang et al. \cite{liang2022effective} conduct the first study to address performance degradation in state-of-the-art self-supervised models for MTL in ADS tasks like semantic segmentation, drivable area segmentation, and traffic object detection. They propose a pre-train-adapt-finetune paradigm that significantly boosts model performance without increasing training overhead. Core to their approach is the LV-Adapter, which incorporates linguistic knowledge from contrastive language--image pre-training (CLIP) by learning task-specific prompts. The experiments highlight the critical role of the adaptation phase in improving MTL, with the language priors from CLIP enhancing performance across multiple downstream tasks. Similarly, VE-Prompt \cite{liang2023visual} utilizes task-specific visual exemplars to guide the model in learning more effective task representations (see Fig. \ref{fig:VE-Prompt}), which alleviates the negative transfer issue among object detection, semantic segmentation, drivable area segmentation, and lane detection. It consists of five parts, including an image encoder, a Transformer encoder, a prompt generator, a task-prompting block, and task-specific heads for different tasks. The prompt generator, utilizing a fixed CLIP image encoder, extracts task-specific prompts from visual exemplars, offering high-quality task-specific knowledge to the model. Furthermore, the framework bridges the Transformer with convolutional layers, enabling efficient and accurate task representation learning. 

Moreover, studies \cite{wei2024bev, tang2025bev, choudhary2024talk2bev} explore the use of VLMs on BEV maps for text-conditioned scene retrieval and interactive BEV question answering in ADS, but they do not adopt MTL. In parallel, Liu et al.\cite{liu2023hierarchical} proposed hierarchical prompt learning for MTL, where tasks are clustered in a tree structure to balance task-shared and task-specific prompts. This approach leverages task-relatedness to capture fine-grained representations. However, it has not yet been evaluated in the context of ADS. Additionally, the applications of language models in ADS have been explored in \cite{wu2023language, chen2024advanced, fu2024drive, jiang2024koma, pan2024vlp}. However, these studies focus on ADS challenges rather than MTL applications.

\begin{table*}[t]
    \centering
    \caption{The results of various MTL methods evaluated on BDD100K.}
    \footnotesize
    \begin{tabular}{@{}p{0.15\textwidth} p{0.04\textwidth} p{0.1\textwidth} p{0.08\textwidth} p{0.07\textwidth} p{0.07\textwidth} p{0.07\textwidth} p{0.07\textwidth} p{0.05\textwidth} p{0.08\textwidth}@{}}
    \toprule
        Methods & Year & Category & mAP50 (\%) & Recall (\%) & mIoU (\%) & IoU (\%) & ACC (\%) & P (M) & FPS\textgreater30 \\ 
    \midrule
        MultiNet \cite{teichmann2018multinet} & 2018 & C-CNN & 60.20 & 81.30 & 71.60 & - & - & - & No  \\ 
        DRMNet \cite{zhao2023drmnet} & 2023 & C-CNN & 80.00 & 93.90 & 92.20 & 27.00 & 76.30 & 8.09 & Yes  \\ 
        UMT-NET \cite{chen2024umt} & 2024 & C-CNN & 79.60 & 92.40 & 94.10 & 28.50 & 77.10 & - & Yes \\
        MEMLModel \cite{miraliev2024real} & 2024 & C-CNN & 77.50 & 86.10 & 91.90 & 33.80 & 76.90 & 6.52 & Yes  \\ 
        MTL R-CNN \cite{rinchen2023scalable} & 2023 & R-CNN & 78.20 & - & - & - & - & - & - \\
        IDS-MODEL \cite{luo2024ids} & 2024 & R-CNN & - & - & 83.63 & - & - & 62.47 & Yes  \\
        JSUMBDP \cite{lee2022joint} & 2022 & YOLO & 76.16 & 89.08 & 92.68 & 26.92 & 72.13 & - & Yes \\
        YOLOP \cite{wu2022yolop} & 2022 & YOLO & 76.50 & 89.20 & 91.50 & 26.20 & 70.50 & 7.90 & Yes  \\ 
        HybridNet \cite{vu2022hybridnets} & 2022 & YOLO & 77.30 & 92.80 & 90.50 & 31.60 & 85.40 & 12.83 & No  \\ 
        YOLO-ODL \cite{guo2023research} & 2023 & YOLO & 79.70 & 94.20 & 92.30 & 27.50 & 75.00 & - & Yes  \\ 
        A-YOLOM(n) \cite{wang2024you} & 2024 & YOLO & 78.00 & 85.30 & 90.50 & 28.20 & 81.30 & 4.43 & Yes  \\ 
        YOLOMH \cite{fang2024yolomh} & 2024 & YOLO & 81.30 & 91.60 & 92.70 & 29.40 & 87.72 & 11.91 & Yes  \\ 
        YOLOPX \cite{zhan2024yolopx} & 2024 & YOLO & 83.30 & 93.70 & 93.20 & 27.20 & 88.60 & 32.90 & Yes  \\ 
        BHF-MTM \cite{zhang2024parallel} & 2024 & YOLO & 84.10 & 91.80 & 93.70 & 31.08 & 89.20 & 35.80 & Yes \\
        MtpNet \cite{li2025mtpnet} & 2025 & YOLO & 89.8 & 94.10 & 95.90 & 29.80 & 86.00 & 50.70 & No \\
        Sparse U-PDP \cite{wang2023sparse} & 2023 & HC-Trans & 84.10 & - & 93.00 & 32.00 & - & 24.50 & No  \\ 
        CUTransNet \cite{li2024cutransnet} & 2024 & HC-Trans & 84.60 & 96.30 & 91.60 & 32.30 & 85.70 & - & Yes  \\ 
        LV-Adapter \cite{liang2022effective}  & 2022 & VLM & - & - & 84.90 & - & - & - & -  \\ 
        VE-Prompt \cite{liang2023visual} & 2023 & VLM & 64.9 & - & 89.40 & 24.00 & - & - & -  \\ 
    \bottomrule
    \end{tabular}

    \vspace{2pt}
    \footnotesize
    \raggedright
    \textit{Note:} \textbf{P}: Number of parameters (in millions). ``-'' indicates not applicable.
    \label{tab:performance_bdd100k}
\end{table*}

\subsubsection{Experimental result}
To facilitate a unified comparison, we collect published results from representative ego vehicle-only MTL models evaluated on BDD100K \cite{yu2020bdd100k} and summarize them in Table \ref{tab:performance_bdd100k}. The compared methods include C-CNN-based models, R-CNN-based models, YOLO-based models, hybrid CNN-Transformer models, and VLM-based models. Note that we do not report object-detection metrics for LV-Adapter because its detection head and data split are not directly comparable to the other BDD100K multi-task settings in Table \ref{tab:performance_bdd100k}. We adopt widely used evaluation metrics for perception tasks in the BDD100K dataset. Specifically, for object detection, we report mAP$_{50}$ and recall. For drivable area segmentation, we use mean Intersection over Union (mIoU). The IoU and ACC columns correspond to lane line segmentation IoU and lane line pixel accuracy, respectively. ``FPS$>$30'' indicates whether the method is reported to achieve real-time inference in the original paper.
It is worth noting that most studies report FPS using different GPU platforms and software environments. Consequently, the real-time performance reported in Table \ref{tab:performance_bdd100k} should be interpreted as indicative rather than strictly comparable across methods.

\begin{table*}[t]
\centering
\caption{Overview of V2X-enhanced cooperative MTL for perception works.}
\label{tab:v2x_perception_compare_3}
\footnotesize
\setlength{\tabcolsep}{3pt}
\renewcommand{\arraystretch}{1.18}
\resizebox{\textwidth}{!}{
\begin{tabular}{l c P{0.10\textwidth} c P{0.10\textwidth} P{0.10\textwidth} p{0.38\textwidth}}
\toprule
Ref. & Year & Main challenges & V2X & Dataset(s) & Tasks & Key advantage \\
\midrule
\cite{tan2024dynamic} & 2024 &
Occlusion / dynamic-scene shifts at intersections &
V2V/V2I &
V2X-Sim &
Det; scene segmentation &
Enhances dynamic-intersection cooperative perception via RSU-guided BEV collaboration and prompt-replay continual adaptation, improving cross-scene robustness under limited RSU storage. \\
\addlinespace
\cite{luo2024multi} & 2024 &
Localization / pose misalignment &
V2V &
Extended OPV2V &
Det; SemSeg; road detection &
Mitigates pose misalignment via cross-agent consensus-based feature correction and attention-based multi-scale fusion, improving multi-task prediction reliability. \\
\addlinespace
\cite{yan2025multi} & 2025 &
Impaired V2X links &
V2V/V2I &
OPV2V; V2XSet &
Det; BEV map segmentation &
Maintains robust cooperative perception under impaired links via two-stage feature recovery and confidence-guided BEV fusion, improving stability of shared BEV representations. \\
\bottomrule
\end{tabular}}
\end{table*}

Based on these metrics, the results show that conventional CNN (C-CNN) and YOLO-based one-stage designs still dominate the accuracy-efficiency trade-off in practice. Recent YOLO-based methods, such as YOLOPX \cite{zhan2024yolopx}, YOLOMH \cite{fang2024yolomh}, BHF-MTM \cite{zhang2024parallel}, A-YOLOM(n) \cite{wang2024you}, and MtpNet \cite{li2025mtpnet}, achieve strong detection and segmentation performance with a wide range of model capacities (about 4-51M parameters in Table \ref{tab:performance_bdd100k}). In particular, many YOLO-family models maintain real-time inference while delivering competitive multi-task accuracy. Although larger variants of YOLO-based methods can further improve accuracy at a higher computational cost, such as MtpNet \cite{li2025mtpnet}, it is hard for them to achieve real-time performance. In contrast, two-stage R-CNN variants either lack full multi-task metrics or incur substantial computational overhead, making them less attractive for onboard deployment. Due to the difficulty of SSD-based methods in detecting small and distant objects in driving scenes, their applicability in recent ego vehicle-only perception MTL has become limited. To the best of our knowledge, no SSD-based MTL models report results on BDD100K under a standard benchmark setting \cite{yu2020bdd100k}. \cite{liu2021mtnas} evaluates the model with a mixed Waymo + BDD100K or CityScapes + BDD100K setup, whose evaluation protocol is not directly comparable. Therefore, SSD-based methods are not included in Table \ref{tab:performance_bdd100k}.

Hybrid CNN-Transformer models occupy a middle ground. Methods such as Sparse U-PDP \cite{wang2023sparse} and CUTransNet \cite{li2024cutransnet} demonstrate improved detection and segmentation accuracy by incorporating global context modeling. However, these gains often come with heavier backbones, and in some cases achieving real-time inference remains challenging.

VLM-based methods, such as LV-Adapter \cite{liang2022effective} and VE-Prompt \cite{liang2023visual}, inject CLIP-based textual or visual prompts into unified perception backbones and have been evaluated on datasets such as BDD100K. Nevertheless, based on the aggregated results, their dense prediction performance still lags behind C-CNN-, YOLO-, and hybrid CNN-Transformer-based MTL models. In addition, current evaluations typically do not report model size or inference latency, limiting their interpretability for onboard deployment. As a result, these approaches are more appropriately viewed as proof-of-concept demonstrations of VLM-guided MTL rather than methods optimized for real-time onboard use.

\subsubsection{Discussion}
As discussed above, ego vehicle-only perception MTL has progressed from early CNN-based architectures toward Transformer, hybrid CNN-Transformer, and VLM-based models that exploit richer context and inter-task relationships. Among these paradigms, CNN-based architectures remain the most widely adopted for onboard deployment, offering a favorable latency-accuracy trade-off due to strong locality and translation equivariance, as well as mature optimization and acceleration support. However, task interaction in CNN-based models is often implicit, relying primarily on hard-parameter sharing with limited explicit modeling of inter-task dependencies.

Transformer-based architectures enhance global context modeling and enable more explicit decoder-level interaction, which is particularly beneficial for unified multi-task decoding and BEV-centric perception. Transformers integrate attention mechanisms into the backbone or BEV fusion level through shared vision Transformer-style encoders and multi-head decoders to support richer multi-task outputs. These advantages, however, often come at the cost of higher computational overhead and stricter training stability requirements, which can complicate real-time onboard deployment. 

Hybrid CNN–Transformer models alleviate the limitations of both CNN- and Transformer-based architectures by retaining convolutional backbones for efficient low-level feature extraction while inserting Transformer blocks to model global context and inter-task interactions. This design supports more unified multi-task decoders while maintaining a more favorable efficiency–accuracy balance than pure Transformer designs.

VLM-based architectures further introduce language or exemplar priors that can enhance transferability and task conditioning, yet existing studies remain at an early stage for onboard use due to limited and non-standardized reporting of model size, memory footprint, and inference latency.

\begin{figure*}[t]
    \centering
    \includegraphics[width=\linewidth]{image/ICA-V2X.pdf}
    \caption{Representative V2X-enhanced cooperative perception MTL architecture (ICA-V2X). ``Cav'' denotes the adjacent collaborative autonomous vehicle, ``ego'' denotes the ego vehicle, and “infra” denotes the infrastructure (adapted from \cite{yan2025multi}).}
    \label{fig:icav2x}
\end{figure*}

\subsection{V2X-Enhanced Cooperative Perception} 
V2X-enhanced cooperative perception extends ego vehicle-only perception to multi-agent scenarios, where vehicles and roadside units (RSUs) exchange complementary information to overcome occlusions, limited field of view, and localization errors. Instead of executing all perception tasks on a single ego vehicle, these methods use V2V and V2I communication to construct shared BEV representations, on top of which multiple task heads are deployed. Compared with ego vehicle-only MTL, V2X-enhanced cooperative perception must additionally handle cross-agent localization errors, sensor heterogeneity, temporal asynchrony, and impaired V2X communication links \cite{caillot2022survey}. These factors impose additional constraints on the design of shared backbones and multi-task heads. Table \ref{tab:v2x_perception_compare_3} summarizes representative V2X-enhanced cooperative perception MTL works, highlighting their challenges, V2X topologies, and advantages.

Current research largely converges on a BEV-centric architecture that couples 3D object detection with semantic segmentation as the dominant cross-task pairing in cooperative perception. In road-to-vehicle frameworks such as AR2VP \cite{tan2024dynamic}, RSUs act as high-capacity sensing hubs. LiDAR point clouds from infrastructure and vehicles are transformed into a common BEV space, processed by a shared backbone typically deployed on the vehicle side, with optional RSU-side offloading when additional compute resources are available, and then fed into a shared multi-task head for 3D detection and scene-level segmentation. To address impaired V2X links, ICA-V2X \cite{yan2025multi} provides a communication-aware cooperative perception pipeline (see Fig. \ref{fig:icav2x}). In this framework, intermediate BEV features are first recovered by a two-stage impaired communication recovery (ICR) module and then fused using a global confidence-guided, ego-centric attention mechanism (V2X-GCF module). A unified multi-task head produces 3D detection and BEV map segmentation outputs. More generally, the segmentation task serves not only as an additional perception output but also as dense auxiliary supervision that regularizes shared BEV representations. This improves robustness under occlusions and missing or degraded messages.

Beyond communication robustness, another emerging trend is to use multi-task heads to enforce cross-agent consensus rather than only fusing raw features. For example, Luo et al. \cite{luo2024multi} construct a global consensus feature map from multi-agent observations and attach multiple task-specific decoders, including 3D object detection, semantic road segmentation, and road detection. A consensus-based feature correction module is then leveraged to align feature representations across agents, mitigating localization errors and improving the reliability of multi-task predictions. In this setting, the additional segmentation and mapping heads provide rich semantic context that facilitates the detection and resolution of inconsistencies, demonstrating that multi-task outputs can be leveraged not only to expand task coverage but also to enhance agreement and robustness in multi-agent perception.

Overall, V2X-enhanced cooperative perception MTL remains confined to a relatively narrow set of BEV-centric task combinations, primarily 3D detection plus semantic or map segmentation \cite{tan2024dynamic,yan2025multi,luo2024multi}. Nevertheless, existing work illustrates how MTL and V2X cooperation can be tightly integrated. The auxiliary segmentation or mapping heads act as structural regularizers for shared BEV features, increase robustness to occlusions and communication impairments, and support consensus-building across agents. Extending these ideas to richer cooperative task sets (e.g., occupancy prediction, instance or panoptic segmentation) and to more realistic communication constraints and sensor heterogeneity is an important direction for future cooperative perception systems.

\subsection{Lessons Learned} 
Ego vehicle-only and V2X-enhanced cooperative perception are constrained by fundamentally different challenges. Ego vehicle-only perception MTL is primarily limited by strict real-time environments and computational budgets, which explains the dominance of one-stage CNN-based backbones and lightweight multi-head designs in practice. In contrast, V2X-enhanced cooperative perception must additionally handle cross-agent localization and pose misalignment, temporal asynchrony, sensor heterogeneity, and impaired V2X links, making robust feature recovery and BEV-based fusion essential. Across both paradigms, sharing a backbone can be insufficient to guarantee effective MTL, as task interactions often remain implicit and inter-task dependencies are only weakly modeled. Accordingly, explicit interaction or conflict-mitigation mechanisms such as joint-attention fusion modules, unified Transformer-based decoders, task-oriented channel scaling, and lightweight modules designed to reduce negative transfer are commonly introduced. Moreover, in cooperative perception, dense segmentation and mapping heads can act as useful regularizers. Beyond providing additional perception outputs, segmentation and mapping heads can serve as dense auxiliary supervision to regularize shared BEV representations and improve robustness under occlusions and degraded messages. They can also support cross-agent consensus by providing dense semantic cues that help detect and correct cross-agent inconsistencies during fusion.

\section{MTL for Prediction}
\label{sec:mtl_prediction}
Prediction-oriented MTL focuses on using a single model to perform multiple prediction-related tasks under strict real-time constraints in either the ego vehicle-only paradigm (when V2X is unavailable) or the V2X-enhanced cooperative paradigm. In this section, the primary goal is to forecast the future behavior, trajectories, and intentions of surrounding agents, often building on upstream perception outputs to provide richer contextual information. Compared with perception, prediction requires modeling long-range temporal dependencies and complex multi-agent interactions while still satisfying real-time and safety requirements on the ego vehicle. In the following subsections, we categorize prediction MTL methods into ego vehicle-only prediction models, which rely solely on onboard sensing and computation, and V2X-enhanced cooperative prediction models that exploit shared information across agents. Finally, we summarize key lessons learned.

\begin{table*}[t]
\centering
\caption{Overview of ego vehicle-only MTL for prediction works.}
\label{tab:onboard_prediction_overview}
\footnotesize
\setlength{\tabcolsep}{2.2pt}
\renewcommand{\arraystretch}{1.12}
\resizebox{\textwidth}{!}{
\begin{tabular}{l c c P{0.1\textwidth} P{0.1\textwidth} c c c c c c c c p{0.32\textwidth}}
\toprule
Ref. & Year & Share & Input & Dataset(s) &
TrajPred & BBoxPred & Intent & Interact & Pose & Track & Attr & Det & Key advantage \\
\midrule

\multicolumn{14}{l}{\textbf{Single-modal}} \\
\midrule
\cite{razali2021pedestrian} & 2021 & H &
RGB images &
(JAAD\,+\,COCO keypoints) &
 &  & \checkmark &  & \checkmark &  &  & \checkmark &
Improves pedestrian crossing-intent prediction via RGB-only multi-task learning with auxiliary pose supervision, maintaining real-time performance in crowded/occluded scenes. \\
\addlinespace
\cite{meng2023lane} & 2023 & Hy &
Vehicle trajectory data &
Driving simulator (self-collected) &
\checkmark &  &  &  &  &  & \checkmark &  &
Lane-change forecasting via hybrid longitudinal–lateral sharing, regularized by trajectory-style supervision. \\
\addlinespace
\cite{yuan2024temporal} & 2024 & Hy &
Vehicle trajectory data &
CitySim &
\checkmark &  & \checkmark &  &  &  &  &  &
Improves long-horizon trajectory prediction by jointly forecasting trajectories and driving intention via gated mixture-of-experts with uncertainty-aware task balancing. \\
\addlinespace
\cite{zhou2024towards} & 2024 & Hy &
Multi-view RGB images &
nuScenes; JAAD; Halpe-Fullbody &
 &  & \checkmark &  & \checkmark & \checkmark & \checkmark & \checkmark &
Reduces redundant computation via a two-stage design (low-res BEV det/3D tracking, then high-res patch attributes), enabling learning with partially labeled datasets. \\
\addlinespace
\cite{yang2024multi} & 2024 & H &
Vehicle trajectory data &
ETH/UCY; ApolloScape &
\checkmark &  &  & \checkmark &  &  &  &  &
Enables explicit interaction modeling by jointly predicting trajectories and interaction indicators via a collision-aware Graph Transformer, stabilizing long-horizon forecasting in dense traffic. \\
\midrule
\multicolumn{14}{l}{\textbf{Multi-modal}} \\
\midrule
\cite{li2024real} & 2024 & Hy &
BEV raster map + Agent trajectories &
nuScenes; Lyft &
\checkmark &  &  &  &  &  &  &  &
Supports real-time heterogeneous-agent trajectory prediction via shared BEV raster context encoding and agent-specific branches in a unified single forward pass. \\
\addlinespace
\cite{chen2024pedestrian} & 2024 & S &
Bounding box coordinates + ego-vehicle speed &
PIE; JAAD &
 & \checkmark & \checkmark &  &  &  &  &  &
Achieves competitive intention and future bounding box prediction using only box coordinates and ego speed via a cross-modal Transformer with uncertainty-aware multi-task balancing. \\
\bottomrule
\end{tabular}}

\vspace{2pt}
\footnotesize
\raggedright
\justifying
\textit{Note:} \textbf{TrajPred}: Trajectory forecasting. \textbf{BBoxPred}: Future bounding box prediction. \textbf{Intent}: Intent prediction. \textbf{Interact}: Interaction prediction. \textbf{Pose}: Pose estimation. \textbf{Track}: Tracking. \textbf{Attr}: Attribute recognition (e.g., age/gender/motion direction).
\end{table*}

\subsection{Ego Vehicle-Only Prediction} Ego vehicle-only prediction focuses on forecasting the future behaviors or trajectories of surrounding agents using only the ego vehicle's onboard sensors and computing platforms. From a task-formulation perspective, some studies \cite{meng2023lane, yang2024multi, yuan2024temporal} focus exclusively on prediction-related tasks, directly inferring future motion from historical trajectories or visual observations. Other studies \cite{razali2021pedestrian, chen2024pedestrian} jointly address perception and prediction within a unified multi-task framework, where perception outputs serve as auxiliary inputs or tasks that enrich contextual representations and regularize the predictor. 

Compared with perception tasks, trajectory prediction requires capturing long-range temporal dependencies in agents' motion histories. This has motivated the widespread use of long short-term memory (LSTM) \cite{hochreiter1997long} models as temporal encoders. These are often combined with convolutional backbones, graph neural networks (GNN), or Transformer blocks to model spatial context and inter-agent interactions. We group the reviewed methods into single-modal and multi-modal categories. Table \ref{tab:onboard_prediction_overview} summarizes representative ego vehicle-only prediction MTL methods. Note that some works adopt an MTL-style formulation by introducing auxiliary supervised branches while keeping all outputs trajectory-related. For example, \cite{li2024real} uses a shared context encoder with agent-specific prediction branches to model heterogeneous road users within a unified framework. For consistency, we mark TrajPred as the primary output. In this context, the term ``multi-task" design refers to agent-type-specific prediction branches that jointly model heterogeneous road users, rather than distinct perception or planning tasks.

\subsubsection{Single-modal prediction} models rely on a single input modality, such as historical trajectory sequences or image data, and leverage MTL to jointly predict multiple related outputs, including trajectories, intentions, and interaction indicators.

\begin{figure}[t]
    \centering
    \includegraphics[width=\linewidth]{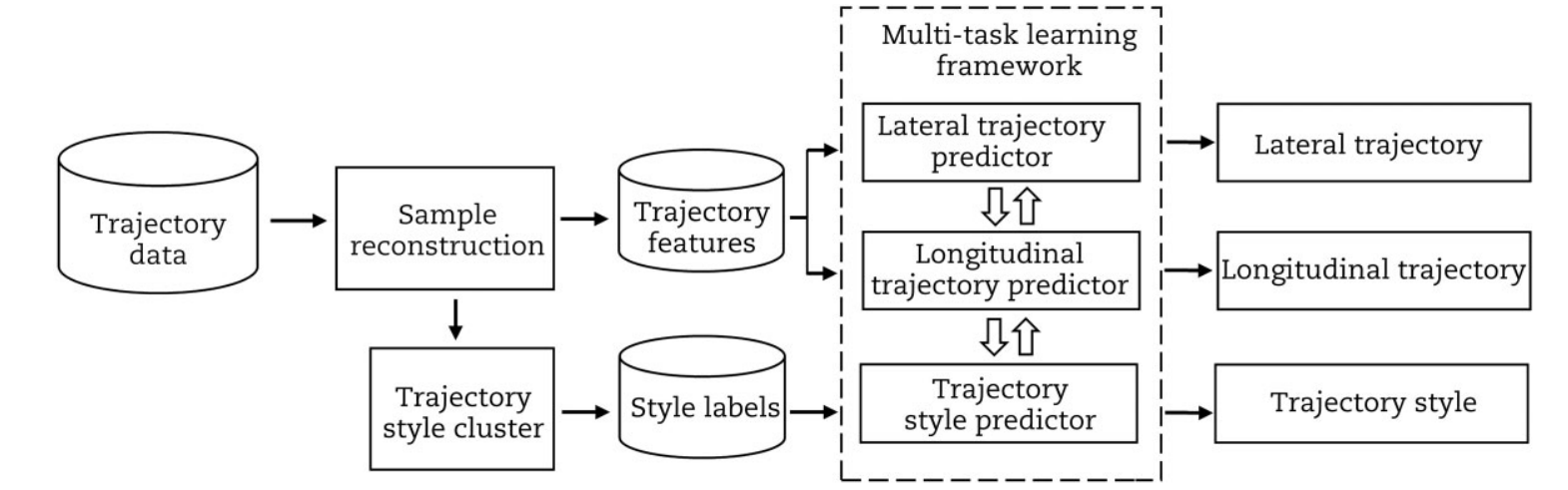}
    \caption{The pipeline of single-modal prediction and hybrid-parameter sharing MTL (adapted from \cite{meng2023lane}).}
    \label{fig:meng2023lane}
\end{figure}

A first line of research builds on trajectory-only inputs and LSTM-style temporal encoders. Hybrid-parameter sharing models \cite{meng2023lane} assign task-specific LSTM branches for lateral and longitudinal trajectory prediction and couple them with a style-recognition head. The resulting features are then fused through an attention mechanism and fed back into each head (Fig. \ref{fig:meng2023lane}). Related designs introduce temporal mixture-of-experts architectures with learnable gating mechanisms \cite{yuan2024temporal}, where shared temporal features are routed to expert subnetworks to jointly predict future trajectories and high-level driving intentions. An uncertainty-aware loss weighting scheme is then adopted to balance the two tasks. 
Studies \cite{yuan2024temporal, yang2024multi} embed historical trajectories and heterogeneous motion states into graph-based or collision-aware attention structures to capture multi-agent interactions. In this framework, auxiliary tasks such as interaction prediction provide additional supervision that stabilizes long-horizon forecasting in dense traffic scenarios.

A second line of single-modal work relies solely on red-green-blue (RGB) camera inputs, using frames to capture pedestrian pose, scene layout, and social context relevant for prediction. In this setting, Razali et al. \cite{razali2021pedestrian} propose a multi-task model that jointly predicts pedestrian crossing intention and pose, where the pose estimation head supplies fine-grained body-configuration features that improve intention classification while maintaining real-time performance. The study in \cite{zhou2024towards} adopts a two-stage efficient design: the first stage performs low-resolution BEV detection and 3D tracking, while the second stage processes cropped high-resolution pedestrian patches for multi-attribute prediction, making the frameworks compatible with partially-labeled datasets. These works illustrate that even without explicit trajectory inputs, visual-only multi-task prediction can leverage semantic diversity to improve robustness and prediction reliability.

\subsubsection{Multi-modal prediction} combines complementary input sources to overcome the limitations of single-modality models. Early approaches rely on rich BEV or rasterized map representations to jointly encode static map context and dynamic agent interactions. For example, the study in \cite{li2024real} rasterizes heterogeneous road information into multi-channel BEV grids and applies hierarchical spatio-temporal encoders with attention-based fusion to model agent motion, social interactions, and map constraints. While such designs provide strong contextual reasoning and trajectory diversity, they depend on dense map availability and incur non-trivial preprocessing and computational overhead, which may limit their applicability in lightweight or real-time scenarios.

To alleviate this limitation, more recent studies explore compact cross-modal prediction designs that fuse minimal yet informative signals without constructing dense BEV or map representations. For example, PedCMT \cite{chen2024pedestrian} combines pedestrian bounding-box trajectories with ego-vehicle speed using a cross-modal Transformer with soft parameter sharing (see Fig. \ref{fig:chen2024pedestrian}). Despite relying on only two lightweight modalities, the model jointly predicts future bounding boxes and crossing intentions using uncertainty-aware multi-task learning. This strategy achieves competitive performance relative to BEV-based approaches. These results suggest that multi-modal MTL can remain effective even under constrained sensing and computation budgets by selectively incorporating complementary cues rather than full scene reconstructions.

\begin{figure}[t]
    \centering
    \includegraphics[width=\linewidth]{image/PedCMT.pdf}
    \caption{Representative multi-modal prediction and soft-parameter sharing MTL architecture (PedCMT) (adapted from \cite{chen2024pedestrian}).}
    \label{fig:chen2024pedestrian}
\end{figure}

\begin{table*}[t]

\centering
\caption{Overview of V2X-enhanced cooperative prediction works.}
\label{tab:v2x_prediction_pp}
\footnotesize
\setlength{\tabcolsep}{2.6pt}
\renewcommand{\arraystretch}{1.12}
\resizebox{\textwidth}{!}{
\begin{tabular}{l c P{0.06\textwidth} P{0.12\textwidth} P{0.08\textwidth} P{0.14\textwidth} P{0.10\textwidth} P{0.08\textwidth} P{0.10\textwidth} p{0.30\textwidth}}
\toprule
Ref. & Year & V2X &
What to share & When to share & How to fuse &
Dataset(s) & Perception outputs & Prediction outputs &
Key advantage \\
\midrule
\cite{wang2020v2vnet} & 2020 & V2V &
Compressed intermediate BEV features &
Multi-step &
GNN/message passing aggregation &
V2V-Sim & Det & TrajPred &
Enables joint Det+TrajPred by delay-compensated, spatially-aware GNN aggregation of compressed BEV features under realistic V2V latency/bandwidth. \\
\addlinespace
\cite{li2024unified} & 2024 & V2I &
Unified BEV features & Not specified &
Deformable-attn BEV encoding + temporal BEV fusion &
DeepAccident & Det\,+\,Track & TrajPred + AccPred &
Unifies Det+Track and motion prediction on cooperative BEV via deformable-attn temporal fusion, then derives accident warning through trajectory post-processing. \\
\addlinespace
\cite{zhou2025v2xpnp} & 2025 & V2V/V2I &
Spatio-temporal BEV features &
One-step &
Transformer spatio-temporal fusion + map-BEV attention &
V2XPnP-Seq & Det & TrajPred &
Preserves spatio-temporal context via one-step intermediate BEV sharing (local multi-frame fusion then transmission), retaining history under bounded bandwidth. \\
\addlinespace
\cite{zhou2025turbotrain} & 2025 & V2V/V2I &
Compressed intermediate features &
One-step &
Multi-frame temporal fusion + multi-agent spatial fusion &
V2XPnP-Seq & Det & TrajPred &
Stabilizes end-to-end multi-agent P\&P training via masked-reconstruction pretraining plus conflict-suppressing gradient balancing, improving performance without manual multi-stage schedules. \\
\bottomrule
\end{tabular}}

\vspace{2pt}
\footnotesize
\raggedright
\justifying
\textit{Note:} \textbf{AccPred}: Accident prediction.
\end{table*}

\begin{figure*}[tb]
    \centering
      \subfigure[What to Transmit]{
        \includegraphics[width=0.60\textwidth]{image/v2xpnp_what.pdf}
        \label{fig:v2xsubfigA}
      }
      \hfill
      \subfigure[When to Transmit]{
        \includegraphics[width=0.36\textwidth]{image/v2xpnp_when.pdf}
        \label{fig:v2xsubfigB}
      }

    \caption{Illustration of various V2X fusion strategies for P\&P (adapted from \cite{zhou2025v2xpnp}).}
    \label{fig:v2xpnp}
\end{figure*}

\subsubsection{Discussion}
Ego vehicle-only prediction MTL has evolved from early LSTM-based, trajectory-centric models to architectures that incorporate interaction cues, visual semantics, and limited sensor fusion \cite{meng2023lane, yuan2024temporal, yang2024multi, razali2021pedestrian, li2024real, chen2024pedestrian}. In practice, single-modal designs are still widely adopted, partly because they require fewer sensors and simpler input pipelines. In these models, auxiliary supervision such as lane change trajectory style recognition, driving intention prediction, interaction probability prediction, and pedestrian pose estimation can help regularize long-horizon forecasting and improve robustness \cite{meng2023lane, yuan2024temporal, yang2024multi, razali2021pedestrian}.  

In contrast, multi-modal approaches enable the incorporation of richer contextual information. For example, rasterized BEV maps support heterogeneous road-agent prediction within a single forward pass \cite{li2024real}. Similarly, PedCMT demonstrates that competitive prediction performance can be achieved without dense scene representations by compactly fusing bounding-box trajectories with ego-vehicle speed \cite{chen2024pedestrian}. These results suggest that carefully selected complementary modalities can offer favorable accuracy–efficiency trade-offs.

However, reported performance improvements are obtained under heterogeneous datasets, task definitions, and evaluation protocols, which hinder unified, task-wise comparisons across studies. As a result, most works demonstrate gains only within their own experimental settings rather than under standardized benchmarks. Accordingly, this survey emphasizes qualitative trends and design trade-offs, such as model complexity, robustness, and deployability, over global numerical rankings. This observation highlights the need for standardized benchmarks and more systematic evaluations for ego vehicle-only prediction MTL.

\subsection{V2X-Enhanced Cooperative Prediction}
V2X-enhanced cooperative prediction extends ego vehicle-only prediction MTL to multi-agent scenarios, where vehicles and RSUs exchange information to mitigate occlusions, limited perception range, and blind spots. In most existing work \cite{wang2020v2vnet, zhou2025v2xpnp, zhou2025turbotrain}, prediction is not treated as an isolated task but is embedded within a joint perception and prediction (P\&P) pipeline. In these frameworks, shared spatio-temporal representations support both 3D detection and trajectory forecasting, sometimes augmented with safety-oriented objectives such as accident prediction. Compared with ego vehicle-only prediction, this paradigm introduces additional design challenges, including what information to transmit over V2X links, when to transmit it, and how to balance P\&P objectives across agents. Table \ref{tab:v2x_prediction_pp} summarizes representative V2X-enhanced cooperative prediction works.

A central design choice in addressing these challenges is how to construct and share spatio-temporal representations across agents under communication constraints. In particular, many approaches adopt shared BEV-centric feature representations as the foundation for cooperative perception and prediction. Accordingly, a first family of methods focuses on cooperative P\&P with shared BEV features. V2VNet \cite{wang2020v2vnet} is an early example in which each vehicle encodes its LiDAR point cloud into an intermediate BEV feature map, compresses this map, and broadcasts it to nearby agents. A spatially-aware GNN aggregates received features using relative poses and time delays, and the fused representation is decoded by task-specific heads for 3D detection and trajectory forecasting. 
By transmitting compressed feature maps instead of raw point clouds, this approach enhances both detection and forecasting performance under severe occlusions while remaining compatible with realistic V2V bandwidth constraints. V2XPnP \cite{zhou2025v2xpnp} generalizes this idea to fully spatio-temporal cooperation. Following the axes in Fig. \ref{fig:v2xpnp}, it systematically studies \textit{what to transmit} (early, intermediate, or late features) and \textit{when to transmit} (multi-step versus one-step communication) and shows that an intermediate-fusion, one-step strategy combined with temporal fusion at each agent is particularly effective. A unified Transformer-based encoder with temporal, self-spatial, multi-agent spatial, and map attention modules then produces a shared spatio-temporal BEV representation, which is jointly decoded for 3D detection and trajectory prediction.

While the above methods primarily aim to improve perception and forecasting accuracy through shared BEV representations, a second family of work extends cooperative P\&P toward explicit safety-oriented prediction objectives. DeepAccident \cite{wang2024deepaccident} introduces a large-scale simulated V2X benchmark with diverse collision scenarios and defines an end-to-end motion-and-accident prediction task. The baseline V2XFormer jointly performs 3D detection and trajectory prediction from multi-view camera histories, while accident occurrence is evaluated by post-processing the predicted trajectories rather than through a separate learned head. Building on this benchmark, UniE2EV2X \cite{li2024unified} proposes a unified V2X end-to-end framework that integrates detection, tracking, trajectory prediction, and accident warning within a single cooperative BEV network. A deformable-attention-based fusion module aligns and merges temporal BEV features from vehicles and infrastructure, and downstream heads are trained for perception and trajectory forecasting, while accident indicators are generated via post-processing over the predicted trajectories. Compared with V2X-enhanced cooperative perception MTL, which mainly exploits V2X fusion and multi-task heads to refine shared BEV representations for detection, segmentation, and cross-agent consensus, this safety-oriented line of work recenters cooperation on prediction. V2X features are fused primarily to forecast future trajectories and derive accident likelihoods, so evaluation goes beyond perception accuracy to include trajectory and accident prediction metrics that directly reflect safety outcomes. Note that the current version of UniE2EV2X \cite{li2024unified} does not report quantitative experimental results. Therefore, we summarize UniE2EV2X mainly as a unified architectural design.

In addition to architectural design, training stability and optimization have emerged as key challenges in cooperative P\&P MTL. The study in \cite{zhou2025turbotrain} observes that simple end-to-end training of multi-agent, multi-frame, multi-task pipelines either collapses or must be replaced by carefully hand-crafted multi-stage schedules. To address this issue, it proposes TurboTrain \cite{zhou2025turbotrain}, which is a two-stage procedure: (i) self-supervised multi-agent spatio-temporal pretraining based on masked reconstruction, which learns task-agnostic 4D representations over agents and time, and (ii) a balanced multi-task optimization stage that suppresses gradient conflicts between cooperative detection and trajectory prediction via a conflict-suppressing gradient-alignment balancer and a hybrid training schedule that alternates between free updates and balanced updates. On the V2XPnP-Seq dataset, this strategy improves both detection AP and joint P\&P metrics while simplifying the training pipeline.

In summary, V2X-enhanced cooperative prediction MTL shows that sharing spatio-temporal BEV representations across agents and jointly optimizing perception and forecasting heads can substantially improve trajectory prediction under occlusions and long-horizon scenarios \cite{wang2020v2vnet, zhou2025v2xpnp, wang2024deepaccident}. At the same time, these studies expose challenges that extend beyond ego vehicle-only prediction, including multi-agent spatio-temporal fusion under bandwidth constraints, the design of safety-oriented auxiliary tasks or post-processing modules such as accident prediction, and the development of training strategies that control gradient interference across agents, tasks, and temporal horizons \cite{wang2020v2vnet, zhou2025v2xpnp, wang2024deepaccident, li2024unified, zhou2025turbotrain}.

\subsection{Lessons Learned}
Prediction-oriented MTL performance is dominated by long-horizon temporal encoders and explicit interaction modeling, rather than task count alone. Auxiliary tasks are most effective when they enrich or regularize these temporal representations rather than act as independent prediction heads. In V2X-enhanced prediction, the accuracy-cost trade-off is largely determined by what information is transmitted, when it is transmitted, and how it is fused. Bandwidth constraints and temporal or pose misalignment frequently bound achievable gains, making communication strategy and fusion policy co-design essential. Moreover, safety-oriented evaluation is critical: conventional trajectory errors may not fully reflect safety outcomes, motivating safety-relevant metrics such as collision or accident risk alongside conventional forecasting measures. Finally, training stability emerges as a first-order concern in multi-agent, multi-task, and long-horizon settings. In practice, staged training or pretraining and conflict-aware optimization often act as enablers for reliable training rather than optional refinements.

\section{MTL for Planning and Control}
\label{sec:mtl_planning_control}
Planning/control-centric MTL in CAVs aims to use a single model to perform multiple planning or control tasks in either the ego vehicle-only paradigm (when V2X is unavailable) or the V2X-enhanced cooperative paradigm. Specifically, planning/control-centric MTL places trajectory generation or low-level control commands at the center of the learning problem and uses auxiliary perception and prediction tasks to shape the shared representation. These methods range from modular pipelines, where planning or control is explicitly separated from perception or prediction, to end-to-end architectures that directly map raw sensor data to waypoints or control commands. In the following subsections, we first examine ego vehicle-only planning/control models that operate on onboard sensing and computing platforms and then move to V2X-enhanced cooperative paradigms in which multiple agents coordinate their planned motions and control policies. Finally, we summarize key lessons learned.

\begin{table*}[t]
\centering
\caption{Overview of ego vehicle-only MTL for planning and control works.}
\label{tab:onboard_planning_control_mtl}
\footnotesize
\setlength{\tabcolsep}{2.4pt}
\renewcommand{\arraystretch}{1.15}
\resizebox{\textwidth}{!}{
\begin{tabular}{l c c P{0.18\textwidth} P{0.08\textwidth} P{0.11\textwidth} P{0.12\textwidth} P{0.08\textwidth} p{0.33\textwidth}}
\toprule
Ref. & Year & Share & Input & Dataset(s) & Planning/Control output & Perception outputs & Prediction outputs & Key advantage \\
\midrule

\multicolumn{9}{l}{\textbf{Planning-centric MTL}} \\
\midrule
\cite{hu2023planning} &
2023 &
Hy &
Multi-camera images &
nuScenes &
Waypoints &
Det; Track; Map &
Mot; FOcc &
Enables planning-centric end-to-end waypoint generation via a unified query-based BEV interface, where Mot/FOcc supervision improves safety and reduces compounding errors. \\
\addlinespace

\cite{ye2023fusionad} &
2023 &
Hy &
LiDAR point cloud data + multi-view camera images &
nuScenes &
Waypoints &
Det; Track; Map &
Mot; FOcc &
Improves multi-modal waypoint planning via BEV camera-LiDAR fusion with modality-aware prediction and status-aware planning modules plus collision-aware loss. \\
\addlinespace

\cite{rao2024enhancing} &
2024 &
Hy &
Monocular images + vehicle measurements + navigation target points + commands &
CARLA (Longest06; Town05-Long) &
Waypoints; target speed &
Det; SemSeg; Depth; Map &
- &
Improves monocular planning stability via temporal fusion and a dedicated motion decoder that better separates motion cues from static BEV context for waypoint/target-speed prediction. \\
\addlinespace

\cite{rao2024camera} &
2024 &
H &
Multi-camera images + vehicle speed + target points + command  &
CARLA (Longest06, Town05-Long) &
Waypoints; target speed &
Det; SemSeg; Depth; Map &
- &
Boosts camera-only planning via depth/semantic-aware BEV transformation and intertask-affinity meta optimization that adaptively reweights tasks to mitigate negative transfer. \\
\midrule

\multicolumn{9}{l}{\textbf{Control-centric MTL}} \\
\midrule
\cite{yang2018end} &
2018 &
H &
Front-camera image + feedback speed &
Udacity; SAIC &
Steer; speed &
- &
- &
Improves end-to-end steering and speed control by fusing front-view vision with historical speed feedback in a multi-branch network. \\
\addlinespace

\cite{ishihara2021multi} &
2021 &
H &
Monocular images + measured velocity + high-level command &
CARLA (Town01; Town02) &
Steer; throttle; brake &
SemSeg; Depth; TL &
- &
Improves closed-loop driving and traffic-light compliance via attention-aware MTL that jointly learns SemSeg/Depth/TL alongside control commands. \\
\addlinespace

\cite{agand2024dmfuser} &
2024 &
Hy &
Multiple RGB-Depth cameras images + speed route command &
CARLA &
Throttle; steer; brake; waypoints &
SemSeg; SDCM; TL &
- &
Improves RGB-Depth end-to-end control robustness via attention-based RGB-SDC fusion with single-task-teacher distillation. \\
\addlinespace

\cite{baicang2025multi} &
2025 &
Hy &
Front-camera RGB image + semantic map + depth image; vehicle-state history &
Augmented Udacity; SCANeR &
Steer; speed &
- &
- &
Enables four-step steer/speed prediction by spatiotemporal fusion of RGB+semantic+depth cues with vehicle-state history, improving robustness in complex driving scenarios. \\
\bottomrule
\end{tabular}}

\vspace{2pt}
\footnotesize
\raggedright
\justifying
\textit{Note:}
\textbf{Map}: HD/BEV map.
\textbf{SDCM}: Semantic depth cloud mapping.
\textbf{Mot}: Motion forecasting.
\textbf{FOcc}: Future occupancy forecasting.
\textbf{TL}: Traffic light state.
\end{table*}

\subsection{Ego Vehicle-Only Planning/Control}
Ego vehicle-only planning/control focuses on generating ego future waypoints or direct low-level control commands on the ego vehicle using only onboard sensors and computing platforms. Most studies treat planning or control as the primary objective, while auxiliary tasks such as perception or prediction are jointly optimized to enrich shared representations, mitigate error accumulation, and reduce redundant computation. Other studies follow an end-to-end paradigm that directly maps raw sensory inputs to control commands, using auxiliary tasks mainly to regularize the latent representation. Broadly, existing ego vehicle-only work can be grouped into planning-centric MTL, where the main head outputs trajectories, waypoints, or target speed, and control-centric MTL, where the main heads output steering, throttle, brake, or speed commands. We review each category in the following subsections. Table \ref{tab:onboard_planning_control_mtl} summarizes representative ego vehicle-only MTL methods.

\subsubsection{Planning-centric MTL}
In planning-centric ego vehicle-only MTL, the model explicitly predicts continuous ego trajectories or waypoints, with perception and prediction treated as supporting tasks. A key trend is to design the backbone and query interaction so that auxiliary tasks genuinely improve planning quality rather than simply sharing features.
UniAD \cite{hu2023planning} is an example of a planning-centric unified model that performs perception, prediction, and planning tasks in a single forward pass. The architecture is primarily designed to be planning-centric and explicitly studies how auxiliary prediction tasks affect the planning head. Ablation studies show that jointly using motion forecasting and occupancy prediction improves planning safety, as reflected in reduced collision rates and trajectory errors, and highlight that both prediction tasks are required for a safe planning objective. The analysis mainly emphasizes their combined effect on planning, rather than deeply dissecting the separate impact of motion versus occupancy. Additionally, the Transformer-based query design facilitates shared attention-based interfaces across tasks, maintaining perception performance while enhancing planning.

Follow-up work explores multi-modal BEV planning and lightweight designs under the same planning-centric philosophy. FusionAD \cite{ye2023fusionad} extends UniAD-style \cite{hu2023planning} ideas to a camera+LiDAR setting: a Transformer-based encoder fuses modalities into a shared BEV space, and modality-aware prediction and status-aware planning modules inject sensor-specific cues and ego state into the planning head. A differentiable collision loss further aligns the learned trajectories with safety objectives. Rao et al. \cite{rao2024camera} investigate a more lightweight camera-only architecture where waypoint planning and target speed prediction are the primary head and depth estimation, semantic segmentation, BEV map generation, and BEV object detection are auxiliary tasks. To reduce task interference, they introduce a meta-learning-based loss weighting scheme that adapts task weights based on inter-task affinity. A complementary work by the same authors \cite{rao2024enhancing} further introduces temporal fusion and a dedicated motion decoder to disentangle dynamic motion features from static BEV features. In summary, these models reveal a common design pattern: 
\begin{itemize}
    \item Planning is positioned at the top of a multi-head hierarchy.
    \item Auxiliary P\&P heads are engineered to provide complementary temporal and semantic structure that supports safer and stable trajectory generation.
\end{itemize}

\subsubsection{Control-centric MTL}
directly targets low-level vehicle commands, such as steering angles and longitudinal speeds. Existing methods mainly follow two families: modular pipelines, where P\&P modules are explicitly separated from control, and end-to-end controllers, where raw sensor inputs are mapped to control signals with auxiliary tasks used only during training.

\begin{figure}[t]
    \centering
    \includegraphics[width=\linewidth]{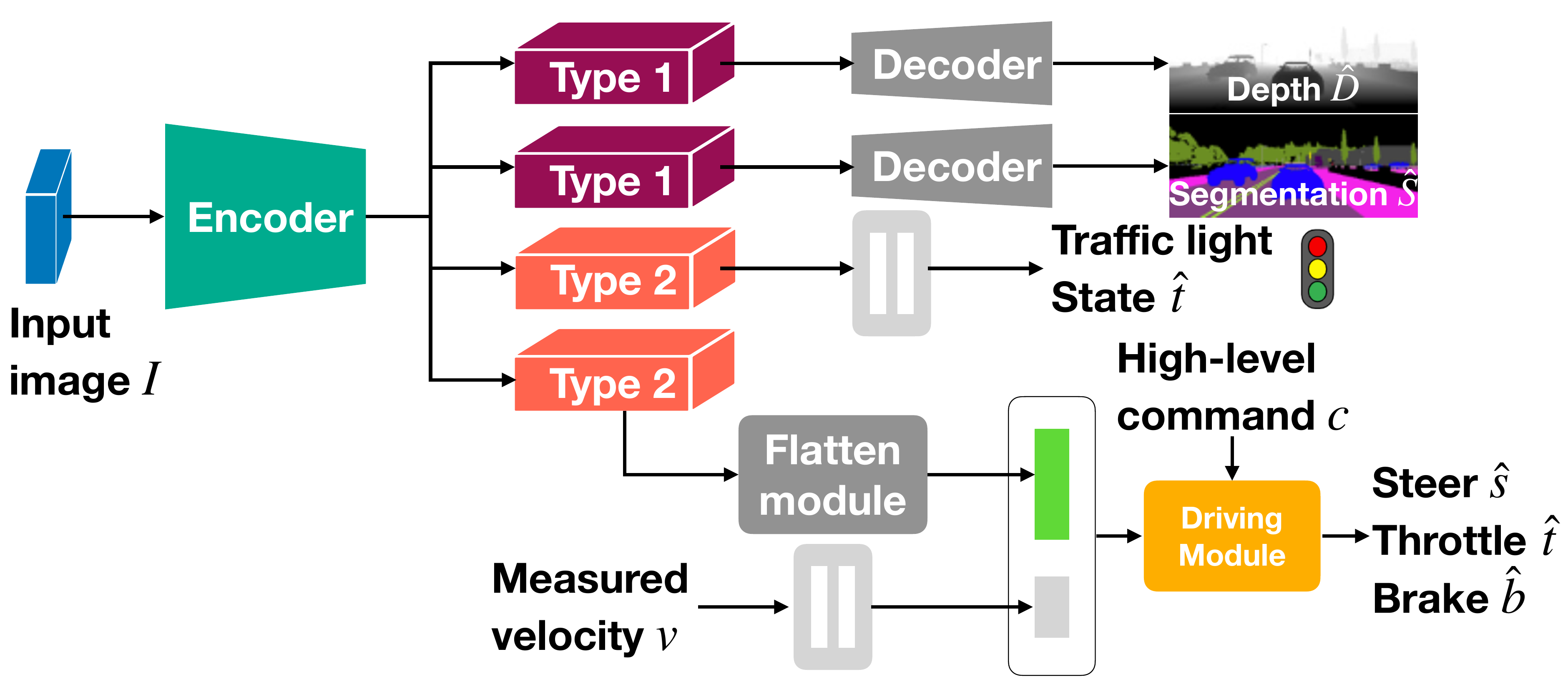}
    \caption{Representative control-centric MTL architecture (adapted from \cite{ishihara2021multi}).}
    \label{fig:ishihara2021multi}
\end{figure}

Modular approaches retain an interpretable perception pipeline while still benefiting from multi-task training. DMFuser \cite{agand2024dmfuser} is a representative that takes RGB-depth inputs and fuses them with semantic depth cloud features using an attention-based CNN module, and jointly learns perception features and control outputs. During training, four single-task teacher networks supervise a multi-task student model through adaptive feature-level distillation, improving representation quality without increasing inference cost. On top of the shared representation, a control module composed of GRU-based branches predicts both trajectory waypoints and low-level vehicle control commands, including steering, throttle, and brake. Similarly, Ishihara et al. \cite{ishihara2021multi}, as shown in Fig. \ref{fig:ishihara2021multi}, propose a multi-task attention-aware network that shares a ResNet-34 backbone and attaches separate heads for depth estimation, semantic segmentation, traffic light classification, and driving control (steer, throttle, brake). Although cross-task interaction is limited to shared features and attention, experiments show that the multi-task attention-aware model with traffic light classification achieves higher driving success rates and fewer red-light violations than single-task control baselines.

End-to-end control-centric MTL methods push task coupling further by directly predicting steering and speed from raw inputs, with MTL used to stabilize training and improve generalization. Early work by Yang et al. \cite{yang2018end} uses front-view images and historical speed feedback as inputs and jointly predicts steering angle and speed in a simple multi-branch network. Even in this minimal setting, MTL significantly improves steering accuracy over single-task baselines, suggesting that jointly learning lateral and longitudinal control can regularize the shared representation. More recent designs enrich both the input modalities and the temporal modeling. Guo et al. \cite{baicang2025multi} propose a multi-modal, multi-task end-to-end controller that ingests RGB images, depth maps, semantic segmentation maps, and historical vehicle states, and jointly predicts future steering and speed. A spatial fusion module combining a channel attention mechanism and a Vision Transformer, together with a Residual-CNN-BiGRU temporal backbone, enables the network to capture complex spatio-temporal dependencies. Multi-step prediction heads further stabilize control over longer horizons and in complex scenarios such as turns, shadows, and scenes with multiple traffic participants.

Overall, modular pipelines preserve interpretable perception outputs and manually designed interfaces between perception and control, whereas end-to-end controllers more tightly couple steering and speed heads to shared spatio-temporal encoders, reducing hand-crafted structure at the cost of interpretability.

\begin{table*}[t]
\centering
\caption{Overview of V2X-enhanced cooperative planning/control works.}
\label{tab:v2x_planning_control}
\footnotesize
\setlength{\tabcolsep}{2.2pt}
\renewcommand{\arraystretch}{1.12}
\resizebox{\textwidth}{!}{
\begin{tabular}{l c c P{0.10\textwidth} P{0.06\textwidth} P{0.15\textwidth} P{0.08\textwidth} P{0.08\textwidth} P{0.08\textwidth} P{0.10\textwidth} p{0.30\textwidth}}
\toprule
Ref. & Year & V2X &
What to share & When to share & How to fuse &
Dataset(s) & Perception outputs & Prediction outputs & Plan/Ctrl output &
Key advantage \\
\midrule
\multicolumn{11}{l}{\textbf{Planning-centric MTL}} \\
\midrule
\cite{yu2025end} &
2025 &
V2I &
Agent/lane queries + occupancy probability map &
One-step &
Flow-compensated, pose-aligned query fusion (Hungarian+MLP); occupancy pose-warp + max fusion &
DAIR-V2X &
Det; Track; Map &
Mot; FOcc &
Waypoints &
Enables planning-centric cooperative driving via sparse–dense intermediate sharing (agent/lane queries + occupancy maps) in a unified query-based BEV framework, improving collision/safety metrics over no-fusion and feature-fusion baselines. \\

\midrule
\multicolumn{11}{l}{\textbf{Control-centric MTL}} \\
\midrule
\cite{huch2021multi} &
2021 &
V2V &
Ego state of leader (velocity, acceleration) &
Multi-step &
State-level fusion via rule-based car-following model &
CARLA (Town01; Town02) &
Traffic-light state &
- &
Leader: steer; throttle; brake. Follower: steer; gap between vehicles &
Enables role-adaptive platoon control via a shared CNN with leader/follower heads, where V2V leader kinematics and a rule-based car-following module stabilize following under camera-only sensing and tight update deadlines. \\
\addlinespace

\cite{lu2025multi} &
2025 &
V2I &
Semantic BEV + local model parameters (for federated aggregation) &
- &
Federated aggregation at RSU + screening-based weighted model fusion &
OpenDD; CARLA &
Object states; Depth; SemSeg; Semantic BEV (grid) &
- &
Steer; throttle; brake; auxiliary waypoint prediction &
Enables risk-aware V2I roundabout control via screening-based dual-aggregation federated RL (FTD3) that fuses RSU semantic BEV with ego vehicle multi-task perception while reducing overhead by federated parameter exchange instead of raw data sharing. \\
\addlinespace

\cite{balasubramanian2025federated} &
2025 &
V2V/V2N &
Local model updates (parameters/gradients) &
- &
Federated model aggregation + GNN message passing &
- &
- & 
Vehicle state forecast
&
Acceleration; steer
&
Enables fleet-level cooperative decision-making via federated multi-task meta-learning that personalizes GNN+MPC state-forecasting controllers without raw data sharing under edge--fog--cloud latency and heterogeneity constraints.  \\
\bottomrule
\end{tabular}}
\end{table*}

\subsubsection{Discussion}
Ego vehicle-only planning/control MTL shows how planning-focused or control-focused heads can be strengthened by auxiliary perception and prediction tasks through shared representations and multi-objective training. Planning-centric systems place trajectory generation at the core and use multi-task supervision to shape BEV or query-based intermediate spaces \cite{hu2023planning, ye2023fusionad, rao2024camera, rao2024enhancing}. In contrast, control-centric methods span modular pipelines and end-to-end controllers \cite{agand2024dmfuser, ishihara2021multi, yang2018end, baicang2025multi}, where sharing features across perception cues and actuation outputs can improve robustness and closed-loop driving outcomes in some reported scenarios. These gains typically come with tighter temporal coupling and additional architectural components, and they are often evaluated under different testbeds and protocols (offline benchmarks versus closed-loop simulation), which limits direct numerical comparison.

\subsection{V2X-Enhanced Cooperative Planning/Control}
V2X-enhanced cooperative planning/control extends ego vehicle-only planning/control to multi-agent scenarios. Depending on the setting, vehicles and infrastructure nodes may exchange inference-time perception/state information for cooperative fusion, or training-time model updates (e.g., parameters/gradients) for federated learning/RL to improve the learned planner/controller. Compared to ego vehicle-only designs, these methods must jointly reason over multi-agent geometry, communication delays, and heterogeneous sensing capabilities, while still keeping planning or control as the primary objective. Existing work can be grouped into \textit{(i)} planning-centric V2X MTL, \textit{(ii)} platoon-level cooperative control, and \textit{(iii)} federated multi-agent control with fleet-level coordination. Table \ref{tab:v2x_planning_control} summarizes representative V2X-enhanced cooperative planning/control works.

Planning-centric V2X MTL frameworks retain the ``planning as primary task'' philosophy of ego vehicle-only models but move to a vehicle–infrastructure scenario. UniV2X \cite{yu2025end} extends UniAD-style \cite{hu2023planning} end-to-end planning to cooperative driving by unifying four tightly coupled tasks in a query-based architecture, including 3D agent perception (3D detection, tracking, and motion forecasting), online mapping via lane and road-element detection, grid-based occupancy prediction, and an ego-trajectory planning head. Infrastructure sensors provide a complementary BEV representation of the scene, and a sparse–dense hybrid transmission scheme sends instance-level queries together with dense occupancy maps over the V2X link. These intermediate tasks are trained jointly in a planning-centric objective, so improvements in cooperative detection, mapping, and occupancy prediction translate into lower collision rates in the final planned trajectories compared to no fusion and BEV feature-fusion baselines \cite{yu2025end}. In this way, V2X communication is used not just to extend perception range, but to learn a planning-centric shared representation across detection, mapping, occupancy, and planning heads.

At the scale of small groups of vehicles, platoon-level cooperative control models show how multi-task architectures can encode role-dependent behaviors. Huch et al. \cite{huch2021multi} study a CAV platoon in which a shared CNN-based network maps front-view camera images to continuous control outputs, with one prediction head driving the leading vehicle (steering and throttle/brake) and another head driving the following vehicle (steering and inter-vehicle gap). Low-level V2V states from the leader (velocity and acceleration) are then combined with the predicted gap in a downstream car-following controller, so the follower can maintain a short headway under limited visibility. Both tasks share most parameters, and a simple task-switching mechanism selects the appropriate head based on vehicle role. The network also predicts traffic-light state as an auxiliary perception output. Experiments show that sharing the network across the leader and follower and incorporating V2V information for the follower improves route-completion rates and yields more stable follower behavior compared to single-vehicle end-to-end controllers \cite{huch2021multi}.

Finally, the third line of work pushes cooperation to the fleet level by combining MTL with federated or multi-agent RL. FedCAV \cite{balasubramanian2025federated} models cooperative planning and control across a large CAV fleet using an edge–fog–cloud architecture. Each vehicle represents neighboring CAVs as a communication graph and performs GNN message passing over vehicle-state features (e.g., position, speed, direction, and status) to learn interaction-aware state representations and predict future vehicle states. An edge MPC module then uses these predictions to compute acceleration and steering actions, optimizing travel time, fuel consumption, and collision risk. Federated multi-task meta-learning aggregates and personalizes the shared model across heterogeneous clients so that different vehicles and regions can adapt to local traffic patterns without sharing raw data. A three-tier edge–fog–cloud topology is used to trade off latency and computation, yielding sub-second end-to-end response times in the reported case study \cite{balasubramanian2025federated}. Complementing this, Lu et al. \cite{lu2025multi} propose a multi-agent federated RL framework for cooperative vehicle–infrastructure control at unsignalized roundabouts. Their federated twin delayed deep deterministic policy gradient (FTD3) architecture, a federated RL variant of TD3 \cite{fujimoto2018addressing}, decomposes the overall decision-making process into several tightly coupled tasks. Specifically, an RSU-side static processing module reconstructs a semantic BEV map, while onboard dynamic processing modules jointly estimate depth, semantic segmentation, semantic BEV (grid), object states, and future waypoints from monocular images and proprioceptive signals. These perception and planning outputs form a cooperative state matrix for a TD3-based continuous-control policy, and a dual-aggregation federated RL mechanism selectively aggregates actor and critic parameters across vehicles based on risk and safety metrics. Experiments on CARLA and OpenDD \cite{breuer2020opendd} roundabout scenarios and a campus map show higher success rates and lower collision rates than centralized RL and simple federated learning plus RL baselines \cite{lu2025multi}, with reduced communication overhead.

Overall, these works adopt a planning- or control-centric MTL paradigm, where trajectory planning or low-level control constitutes the primary head, while perception and prediction tasks, often enhanced through V2X communication, serve as auxiliary components to improve safety, robustness, and coordination across agents. Planning-centric frameworks such as UniV2X \cite{yu2025end} highlight multi-task query design and hybrid sparse–dense transmission. Platoon-level models show how shared backbones and role-specific heads encode coordinated control policies \cite{huch2021multi}. Federated multi-agent approaches such as FedCAV and FTD3-based control \cite{balasubramanian2025federated, lu2025multi} illustrate how fleet-level cooperation and privacy-preserving training can be layered on top of multi-task representations.

\begin{table*}[t]
\centering
\caption{Overview of MTL for V2X communications and RRM works.}
\label{tab:v2x_mtl_comm_compare}
\footnotesize
\setlength{\tabcolsep}{2.2pt}
\renewcommand{\arraystretch}{1.15}
\resizebox{\textwidth}{!}{
\begin{tabular}{l c c P{0.23\textwidth} P{0.20\textwidth} P{0.24\textwidth} p{0.28\textwidth}}
\toprule
Ref. & Year & V2X &
Scenario &
Multi-task outputs  &
Method  &
Key advantage \\
\midrule
\cite{parvini2023aoi} &
2023 &
V2V+V2I &
Platoon-based C-V2X RRM with joint intra-platoon CAM dissemination and RSU-centric AoI control &
AoI minimization (V2I); CAM delivery success probability (V2V) &
Multi-agent multi-task RL: global critic + per-agent local critic; TD3-style global critic; task-wise reward decomposition and objective-wise value learning &
Jointly optimizes RSU-side information freshness and intra-platoon CAM delivery success probability via distributed multi-agent multi-task RL, avoiding centralized scheduling overhead while improving platoon-level communication quality. \\
\addlinespace

\cite{eldeeb2024multi} &
2024 &
V2N &
CAV-to-CAV semantic traffic-sign sharing via LEO satellite (vehicle–satellite–vehicle) under impaired visibility &
Image reconstruction; image classification &
Multi-task semantic communication: shared convolutional autoencoder-based semantic encoder + task-specific decoders &
Transmits compact semantics to support both reconstruction and classification, achieving higher performance than baselines in low-SNR satellite links while saving up to 89\% bandwidth. \\
\addlinespace

\cite{iliadissignal} &
2025 &
V2N &
Hardware-aware multi-cell C-V2X radio link metrics prediction for edge deployment &
Radio-link metric prediction for primary and secondary cells &
MT-FT-Transformer with shared encoder and task-specific heads, combined with hardware-aware pruning and quantization &
Improves the accuracy–latency trade-off of multi-cell C-V2X link-metric prediction, enabling deployment on resource-constrained vehicular/edge hardware. \\
\bottomrule
\end{tabular}}

\vspace{2pt}
\footnotesize
\raggedright
\justifying
\textit{Note:} \textbf{AoI}: Age of information. \textbf{CAM}: Cooperative awareness message.
\textbf{LEO}: Low Earth orbit.
\end{table*}

\subsection{Lessons Learned}

In planning/control-centric MTL, planning or control losses are often considered as the primary learning objectives. Auxiliary perception and prediction heads then regularize shared BEV or query representations, which tends to improve downstream task metrics such as collisions, infractions, and stability. Modular pipelines preserve explicit interfaces and intermediate outputs. This design facilitates debugging and safety validation. By contrast, end-to-end pipelines tightly integrate perception and control and may exhibit greater robustness, but they are harder to interpret and typically require stronger regularization and more careful closed-loop evaluation. Moreover, fleet-scale cooperation reshapes MTL by introducing data heterogeneity and privacy constraints. Raw-data sharing is usually impractical across vehicles and regions. This motivates federated multi-task, meta-learning, and federated RL approaches. These methods emphasize privacy-preserving parameter exchange, global aggregation, and client personalization under heterogeneous compute and communication budgets.

\section{MTL for V2X Communications and RRM}

Unlike earlier V2X-enhanced cooperative discussions, which primarily treat communication as an enabler for cooperative perception, prediction, planning, and control, this subsection focuses on MTL problems that arise within the V2X communication stack itself and on network-side RRM. In this context, the learning objectives are communication-centric, including communication metrics, semantic payload representations, and radio resource allocation decisions. In this scope, MTL refers to the joint learning of shared representations or shared policies that optimize multiple communication-centric objectives, rather than the orchestration of independent applications at the system layer. Although architectures such as that shown in Fig. \ref{fig:communication} may appear similar to V2X perception frameworks, they are communication-centric in nature, as tasks such as classification and reconstruction serve as semantic decoding objectives to optimize payload efficiency and robustness over V2X links, rather than to support cooperative sensing. Table \ref{tab:v2x_mtl_comm_compare} provides an overview of representative works, summarizing their scenarios, multi-task outputs, methods, and key advantages.

\label{sectionivd}
\begin{figure}[t]
    \centering
    \includegraphics[width=\linewidth]{image/communication.pdf}
    \caption{Representative task-oriented semantic communication MTL (adapted from \cite{eldeeb2024multi}).}
    \label{fig:communication}
\end{figure}

From a design perspective,  existing studies can be categorized into three recurring MTL design patterns. First, multi-output supervised learning treats heterogeneous communication KPIs as correlated labels, enabling a shared backbone to exploit cross-metric structure while maintaining lightweight task heads for deployment \cite{iliadissignal}. Second, multi-task representation learning for semantic communication learns a compact latent representation that is simultaneously sufficient for multiple downstream objectives at the receiver, aligning the transmitted payloads with task utility under bandwidth and SNR constraints \cite{eldeeb2024multi}. Third, multi-objective multi-agent RL formulates RRM as learning shared policies under coupled objectives such as freshness, reliability, and access success, where reward decomposition and global-local critics are commonly used to stabilize learning while preserving each agent's local utility \cite{parvini2023aoi}.

Representative works across different layers of the communication stack instantiate these design patterns. In the physical layer, Iliadis et al. \cite{iliadissignal} build a hardware-aware multi-task FT-Transformer \cite{gorishniy2021revisiting} to jointly predict radio-link metrics for the primary and secondary cells in C-V2X. To meet edge constraints, they further introduce a compression pipeline that co-optimizes accuracy and inference latency via pruning and quantization. Moving from bit-level transmission to goal-driven delivery, Eldeeb et al. \cite{eldeeb2024multi} propose a semantic communication framework in which a shared convolutional autoencoder performs semantic encoding of traffic-sign images (Fig. \ref{fig:communication}). Two task-centric decoders at the receiver support image reconstruction and classification, enabling robust performance under low signal-to-noise ratio (SNR) while reducing transmitted payload. At the network control layer, Parvini et al. \cite{parvini2023aoi} formulate platoon-based C-V2X spectrum access as a multi-agent multi-task RL problem that jointly targets AoI reduction and CAM delivery success probability. Their design combines global-local critics for coordination and reward decomposition for objective-wise value learning.

Despite these promising exemplars, the literature on communication-centric MTL remains sparse compared with MTL applied in ego vehicle software layers. This gap arises from two complementary factors. First, many V2X communication and RRM problems admit strong and interpretable solutions based on optimization and standardized protocol design, making the incremental benefits of deep MTL more difficult to demonstrate. Second, deploying and evaluating communication-centric learning systems entails substantial practical overhead. In particular, realistic large-scale datasets and standardized testbeds for multi-KPI prediction and multi-objective resource control are scarce, and empirical results can be highly sensitive to channel models, mobility patterns, and hardware configurations. Moreover, stringent reliability and latency requirements, together with standard compliance, severely constrain the feasible policy space and increase the cost of validating generalization beyond simulation. Finally, MTL gains must be realized under strict inference latency and compute budgets on edge devices, where compression and hardware-aware co-design are first-order requirements rather than optional enhancements.
As a result, deep MTL improvements are often difficult to justify consistently and to reproduce across platforms and scenarios, naturally leading to a limited number of well-scoped case studies. 

Looking forward, an important direction is to couple the communication-layer MTL with downstream driving utilities more explicitly. For example, future systems may jointly optimize resource allocation and semantic payload design for both network KPIs and safety-critical perception or cooperative awareness objectives, while maintaining verifiable reliability under deployment constraints.

\subsection{Lessons Learned}  
MTL is most effective for communications and RRM when target KPIs share latent structure, such as correlated link-quality metrics across cells, bands, or users. Reported gains are often sensitive to channel models, mobility patterns, and hardware configurations, especially under scenario shifts. In task-oriented semantic communication, the downstream utility under explicit bandwidth and SNR constraints is the primary objective. Preserving receiver-side task performance is more important than reconstruction fidelity alone. For RRM, learning-based gains are most interpretable when compared against strong optimization and protocol-based baselines, as many objectives already admit mature, constraint-satisfying solutions. Consequently, deep MTL improvements must be justified under strict latency, reliability, and deployment constraints to demonstrate practical value.

\section{Discussion and Future Directions}
\label{section5}

\subsection{Discussion}
\label{sec:cross_module_discussion}

In this study, we review MTL in CAVs through a taxonomy defined by two complementary dimensions: functional domain and operational context. Specifically, MTL methods are organized into perception, prediction, planning and control, as well as V2X communications and RRM. The perception, prediction, planning, and control domains are further examined under two operational contexts, namely ego vehicle-only and V2X-enhanced cooperative. In contrast, V2X communications and RRM are considered as a communication-centric problem, since their learning objectives are inherently tied to V2X links and network control loops rather than ego vehicle driving functions. Overall, this taxonomy highlights that MTL in CAVs goes beyond simply bundling multiple tasks from software layers into a single network. It also encompasses communication-centric objectives such as multi-KPI link-quality prediction, semantic communication for task-driven payloads, and multi-objective RRM policy learning \cite{iliadissignal, eldeeb2024multi, parvini2023aoi}.

Across ego vehicle-only perception, prediction, and planning and control, a recurring observation is that simple backbone-level sharing is often insufficient \cite{rao2024enhancing, hu2023planning}. Performance gains are more likely when tasks are coupled through semantically meaningful intermediate representations \cite{wang2023sparse, chen2024umt}, in which auxiliary heads regularize the shared space rather than competing for low-level features. When tasks differ in supervision density and inductive bias (e.g., sparse detection versus dense segmentation), backbone-only sharing can introduce negative interference. This motivates designs that couple tasks through structured intermediate representations, including BEV features or query embeddings \cite{rao2024enhancing, wang2023sparse, hu2023planning}. This trend is evident in planning- and control-centric models, where auxiliary perception or prediction heads are mainly used to support the primary planning or control tasks through richer shared representations, rather than being introduced to optimize the auxiliary tasks in isolation \cite{hu2023planning, ye2023fusionad}.

The ego vehicle-only and the V2X-enhanced cooperative paradigms are constrained by different factors. Ego vehicle-only MTL is primarily limited by vehicle-side latency, compute, and memory budgets, which explains the prevalence of efficient shared backbones with lightweight heads in practice \cite{wu2022yolop, miraliev2024real}. In contrast, V2X-enhanced cooperative MTL is often bounded by collaboration reliability, including bandwidth limitations, packet loss, time misalignment, and pose errors. As a result, recent progress has focused on communication-aware recovery mechanisms and robust multi-agent fusion, rather than simply scaling local backbone capacity \cite{tan2024dynamic, luo2024multi, yan2025multi, wang2020v2vnet, zhou2025v2xpnp}.

Comparing results across studies remains challenging because reported performance is highly sensitive to evaluation settings and underlying assumptions. In perception, accuracy and efficiency are often measured under different hardware and runtime configurations, which complicates the comparison of speed and cost claims \cite{wu2022yolop, miraliev2024real, wang2023sparse}. In prediction and planning/control, outcomes depend strongly on dataset choices and evaluation protocols, including forecasting horizons and whether assessment is conducted offline or in closed-loop settings \cite{meng2023lane, chen2024pedestrian, hu2023planning, rao2024camera}. In cooperative scenarios, reported gains are further influenced by communication and alignment assumptions, such as bandwidth constraints, message loss or feature corruption, latency, and pose errors \cite{caillot2022survey, wang2020v2vnet, yan2025multi}. For V2X communications and RRM, learning-based approaches should also be benchmarked against strong optimization- or protocol-based baselines under explicit latency and reliability constraints, which makes demonstrating consistent and reproducible gains more difficult \cite{parvini2023aoi, iliadissignal}.

These observations motivate the translation of the identified patterns into concrete research gaps and actionable future directions, which are discussed in the following subsection.

\subsection{Research Gaps and Future Directions}
\label{sec:research_gaps_future}

\subsubsection{\textbf{Task formulation and relatedness modeling}}
\label{sec:gap_formulation}

A key research gap lies in how multi-task objectives are formulated and how task-relatedness is modeled. In CAV scenarios, task-relatedness is rarely static and can vary with scenario complexity, sensing conditions, prediction horizon, and, in cooperative setups, link quality and agent heterogeneity. As a result, a static multi-task formulation may perform well in some regimes but lead to negative transfer in others. This issue commonly arises when tasks have mismatched supervision density or inductive bias, or when auxiliary objectives are introduced without clear evidence that they support the primary objective \cite{liang2023visual, rao2024enhancing}. Moreover, many designs rely on task-specific prior knowledge, such as selecting which tasks to combine, determining where to share representations, and deciding which heads to decouple. This requires increased design complexity and limited portability across datasets and platforms \cite{vu2022hybridnets, zhan2024yolopx, li2024real}.

Future work should move toward context-dependent task-relatedness and objective design that aligns with operational evaluation. Key directions include scenario-aware task coupling under varying conditions, principled measures to verify whether auxiliary losses genuinely benefit the main objective, and closed-loop formulations for planning and control that prioritize safety and stability rather than only optimizing offline metrics.

\subsubsection{\textbf{Architecture and parameter-sharing under system constraints}}
\label{sec:gap_arch_system}

Building on Section \ref{sec:cross_module_discussion}, ego vehicle-only MTL is ultimately constrained by real-time deployability rather than offline accuracy alone. A key gap is that architecture design is still rarely driven by these deployment constraints from the outset. Although model compression and runtime acceleration are widely adopted, many studies first optimize accuracy under controlled experimental settings and then apply pruning, quantization, or engine-level acceleration as a post-processing step \cite{teichmann2018multinet}. However, in real deployments, choices regarding sharing patterns, head structures, and intermediate representations must account for hardware characteristics and runtime behavior. This consideration is important for real-time systems, where latency is often dominated by kernel or operator efficiency and memory movement rather than parameter count alone \cite{chang2021yoltrack}.

Future directions should move toward deployment-first MTL design. First, hardware-aware multi-task architecture search is needed to ensure that models are optimized for the target devices rather than only for experimental setups. Second, parameter sharing strategies should be structured to maintain inference efficiency, for example by preserving operator-level efficiency and avoiding designs that introduce excessive memory movement or fragmented execution. Third, compression should be co-designed with task coupling rather than applied as a separate post-processing step, since pruning and quantization can change task interference patterns and accuracy trade-offs. In addition, developing more unified architectures that reduce task-specific branches while still controlling negative transfer remains an important direction. Lightweight modules such as adapters or routing blocks offer a practical compromise, but their effectiveness should be validated through transparent runtime reporting, including hardware, precision, batch size, and latency breakdown, to demonstrate tangible deployment gains.

\subsubsection{\textbf{Evaluation, benchmarks, and reproducibility}}
\label{sec:gap_benchmark}

As discussed in Section \ref{sec:cross_module_discussion}, cross-study conclusions are often sensitive to evaluation protocols and underlying system assumptions. A key gap lies in the absence of standardized, end-to-end benchmarking practices that enable fair and reproducible comparison of multi-task CAV systems. In perception, accuracy and efficiency are reported under different hardware platforms and runtime configurations, making comparisons of speed and computational cost difficult \cite{wu2022yolop, miraliev2024real, wang2023sparse}. In prediction and planning/control, reported performance depends on dataset selection and evaluation protocol, including forecasting horizons and whether evaluation is conducted offline or closed-loop settings \cite{meng2023lane, chen2024pedestrian, hu2023planning, rao2024camera}. Dataset limitations and domain shifts further complicate interpretation, as gains observed on a single benchmark may not generalize across regions, weather conditions, or sensing configurations \cite{fang2023improved, ishihara2021multi}.

Future research should advance along three main directions. First, a minimum reporting checklist for runtime and system cost should be established, including hardware specifications, numerical precision, acceleration frameworks, end-to-end latency breakdown, and memory footprint. Second, benchmark suites should be developed to evaluate joint performance using both task-level metrics and system-level metrics, instead of reporting isolated per-task results. Third, studies should include single-task baselines and report performance gains or degradations when transitioning to multi-task training under the same compute budget and deployment settings, clarifying when MTL is beneficial and when it is detrimental. In addition, benchmarks should incorporate standardized cross-domain generalization tests (e.g., region, weather, and sensor-stack shifts) to assess robustness beyond a single dataset. For planning and control, broader adoption of comparable closed-loop evaluation frameworks would improve reproducibility. 

\subsubsection{\textbf{V2X-enhanced cooperative MTL under imperfect links and heterogeneity}}
\label{sec:gap_v2x_coop}

Based on the discussion in Section \ref{sec:cross_module_discussion}, cooperative MTL is shaped by collaboration reliability and cross-agent alignment rather than by local model capacity alone. A key gap is the lack of robust and general cooperative MTL designs that remain effective under heterogeneous agents and imperfect collaboration conditions. In practice, cooperative systems must exchange and align information across agents under heterogeneous sensors, different viewpoints, and time-varying communication quality. Bandwidth limits, message loss or feature corruption, transmission delays, and pose errors can dominate end-to-end performance and may even alter whether multiple objectives remain mutually beneficial \cite{tan2024dynamic, yan2025multi}. These factors make it difficult to design unified cooperative MTL models that are robust across realistic link conditions and diverse agent configurations.

Future work should treat cooperative MTL as a joint learning-and-communication design problem. Key directions include (i) link- and uncertainty-aware collaboration strategies that decide what information is transmitted and how it is fused under bandwidth and reliability constraints, (ii) intermediate representations that remain effective under imperfect temporal or spatial alignment, and (iii) training and evaluation protocols that explicitly reflect agent heterogeneity in practice rather than assuming identical sensors and synchronized clocks. Overall, sustained progress will likely depend on realistic evaluation and end-to-end system integration, not solely on architectural innovations.

\subsubsection{\textbf{Safety, assurance, and responsible AI for MTL in ADS}}
\label{sec:gap_safety_responsible}

ADS are safety-critical, yet safety assurance is still not a primary design objective in most MTL pipelines. Many studies emphasize average-case performance improvement, whereas real-world deployment requires robust behavior under rare events, adverse weather conditions, and sensor failures \cite{lin2018architectural, chen2024advanced, ceccarelli2022rgb}. In addition, MTL can introduce failure coupling, as shared representations may cause errors in one task to propagate to other tasks, thereby increasing the difficulty of system-level validation and certification.

Future research should strengthen safety assurance in MTL-based ADS along several dimensions. First, evaluation protocols should move beyond standard test splits and explicitly assess deployment risks, including rare events, adverse environmental conditions, and sensor degradations, using safety-oriented metrics aligned with the intended operating conditions. Second, models should incorporate runtime monitoring and safe fallback mechanisms to detect high uncertainty, cross-task inconsistencies, and out-of-distribution inputs, ensuring that errors are contained rather than propagated through shared representations. Third, responsible AI considerations, including bias, transparency, and accountability, should be revisited in MTL, as shared representations can propagate and potentially amplify systematic errors across multiple outputs.

\section{Conclusion}
\label{sec:conclusion}
This survey provides a comprehensive review of MTL methods in CAVs. We first present a systematic overview of CAV systems, spanning from hardware and software layers to V2X communication. Then, we introduce the core MTL concepts, including problem formulation, parameter-sharing paradigms, and optimization techniques. Next, we review recent advances in deep MTL for CAVs from 2018 to 2025, covering perception, prediction, planning, control, and V2X communications and RRM. The first four domains are discussed under ego vehicle-only and V2X-enhanced cooperative paradigms, while V2X communications and RRM are discussed as communication-centric MTL problems. Finally, we identify key research gaps and outline promising future directions for MTL in CAVs. We hope this survey serves as a useful reference for researchers seeking to deepen understanding and advance innovation in this area.

\bibliographystyle{IEEEtran}

\bibliography{bibliography}

\end{document}